%% file: main.tex
\documentclass[letterpaper, 10 pt, conference]{ieeeconf}  % Comment this line out if you need a4paper
\usepackage{graphicx}
\usepackage{subcaption}
\usepackage{lipsum} % For dummy text (remove in your actual document)
\usepackage{amsmath,amsfonts}
\usepackage{comment}

\IEEEoverridecommandlockouts                              % This command is only needed if 
\input{preamble}

\title{\LARGE \bf
One Ring to Rule Them All: Constrained Distributional Control for Massive-Scale Heterogeneous Robotic Ensemble Systems
}

\author{Andres Arias$^{1}$, Wei Zhang$^{2}$, Haoyu Qian$^{2}$, Jr-Shin Li$^{2}$ and Chuangchuang Sun$^{1}$% <-this % stops a space
\thanks{$^{1}$ Andres Arias and Chuangchuang Sun are with the Department of Mechanical Engineering, Villanova University, Villanova, PA, 19085.}%
\thanks{$^{2}$ Wei Zhang, Haoyu Qian and Jr-Shin Li are with the Department of Electrical and Systems Engineering, Washington University in St. Louis, St. Louis, MO 63130}%
}   
\begin{document}

\maketitle
\thispagestyle{empty}
\pagestyle{empty}

%%%%%%%%%%%%%%%%%%%%%%%%%%%%%%%%%%%%%%%%%%%%%%%%%%%%%%%%%%%%%%%%%%%%%%%%%%%%%%%%
\begin{abstract}
Ensemble control aims to steer a population of dynamical systems using a shared control input. This paper introduces a constrained ensemble control framework for parameterized, heterogeneous robotic systems operating under state and environmental constraints, such as obstacle avoidance. We develop a moment kernel transform that maps the parameterized ensemble dynamics to the moment system in a kernel space, enabling the characterization of population-level behavior. The state-space constraints, such as polyhedral waypoints to be visited and obstacles to be avoided, are also transformed into the moment space, leading to a unified formulation for safe, large-scale ensemble control. Expressive signal temporal logic specifications are employed to encode complex visit-avoid tasks, which are achieved through a single shared controller synthesized from our constrained ensemble control formulation. Simulation and hardware experiments demonstrate the effectiveness of the proposed approach in safely and efficiently controlling robotic ensembles within constrained environments.

% Different nonlinear and mixed integer nonlinear approaches are presented throughout the experimentations, considering: box constraint and polyhedron constraint; way-point visiting using Signal Temporal Logic (STL) in the ensemble; and obstacle avoidance with disjunctive equations. Prominent results are obtained in simulation and hardware experimentation to contribute to the ensemble control of robots under constrained environments. 

% \red{too specific/ detailed, like unicyle}

\end{abstract}

%%%%%%%%%%%%%%%%%%%%%%%%%%%%%%%%%%%%%%%%%%%%%%%%%%%%%%%%%%%%%%%%%%%%%%%%%%%%%%%%
\section{INTRODUCTION}

The field of robotics has experienced significant advancements in the development of efficient, scalable, and robust architectures that enable a swarm of robots with a shared dynamics structure to perform complex tasks in a mutual collaboration. In this context, the collection of dynamical systems at a large scale, each of them following a similar structure with heterogeneous dynamics, is defined as an ensemble system. Parameters or initial conditions can determine the difference between each system. 
%Applications of ensemble systems are found in natural processes \cite{yamashkin2018using}, summary language and psychology \cite{malik2024ensemble}, disease prediction \cite{cheng2021integrating}, molecular biology \cite{onuchic1996protein}, and engineering control systems \cite{li2007ensemble}. Some multidisciplinary examples include, but are not limited to, control of large number of electron spins for quantum computing \cite{wesenberg2009quantum}, study of stationary and dynamical properties of neuronal ensembles using a generalized rate model in neuroscience \cite{hasegawa2007generalized}, and modeling thermostatically controlled loads ensembles in power distribution systems highly penetrated by renewable energy \cite{hassan2018optimal}. 
Ensemble systems applications are found in engineering control systems \cite{li2007ensemble}, and some multidisciplinary efforts include quantum computing \cite{wesenberg2009quantum} and power systems \cite{hassan2018optimal}. This work focuses on ensemble systems of autonomous vehicles with parametric variation.
%Consider a heterogeneous swarm of autonomous agents that share the same dynamics structure but differ in their parameter values. These variations encompass features such as design (different frames and sizes of drones), manufacturing (different mass and friction values of internal components), and external environment factors (off-road terrains with different traction and traversability).

\begin{figure}[h]
    \centering
   \includegraphics[width=1\linewidth]{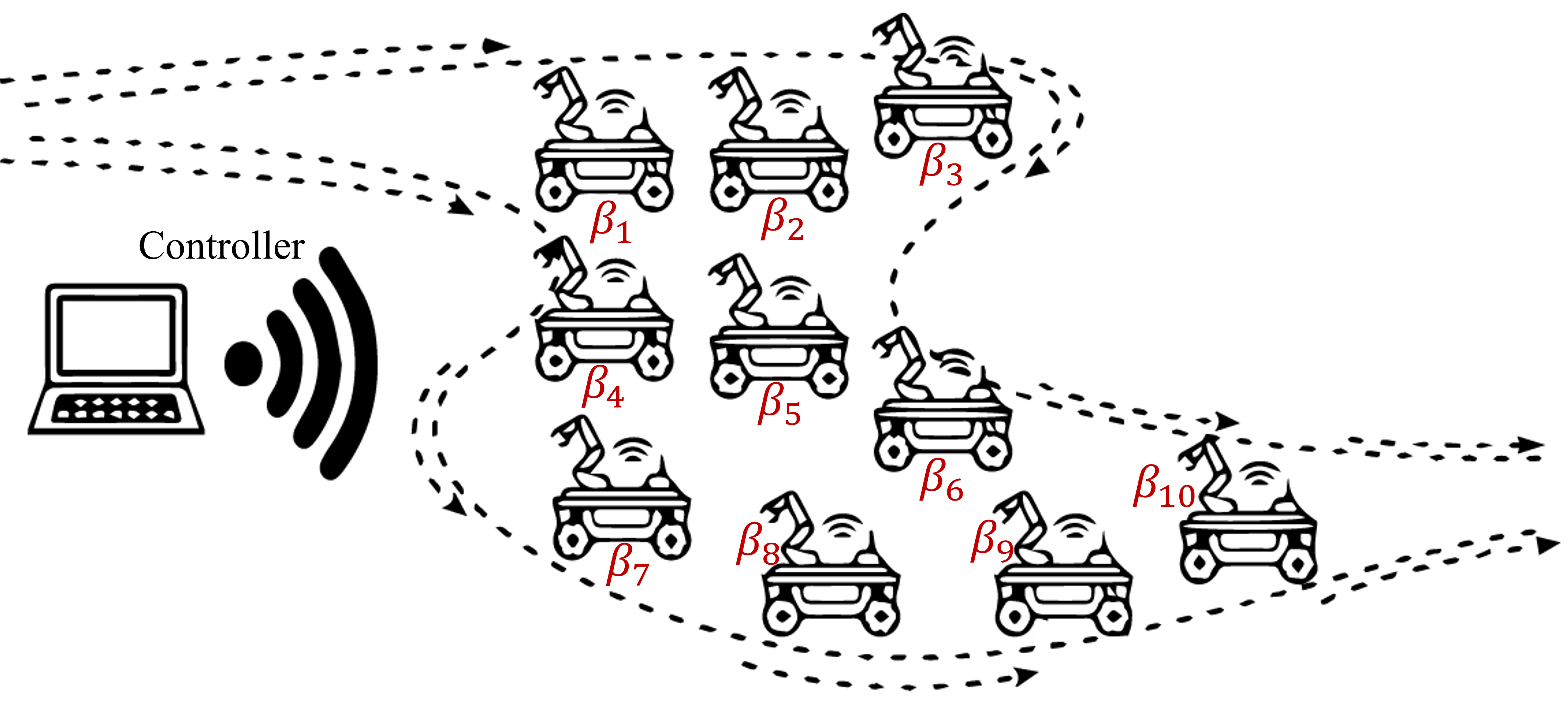}
    \caption{Broadcasting shared control to an ensemble of heterogeneous robots parameterized by $\beta$. 
    % \red{change the color of beta? add "controller" on top of the wifi sign}
    }
    \label{fig:EnsembleBroad}
\end{figure}

In a heterogeneous swarm of autonomous robots, the structure at each unit is identical but with different parameter configurations. Features in these variations include design: different frames and sizes of drones; manufacturing: different mass and friction values of internal components; and external environment factors: off-road terrains with different traction and traversability. From the statistical point of view and considering a large-scale rollout, it is expected that the swarm behaves in an intelligent coordination, following a distribution of location, an organized formation, or a predefined trajectory, which turns into a complex and emergent behavior of the swarm with novel functionalities. Versatile examples of great real-world interest include the deployment of a swarm of quadcopters with different values of mass or moment of inertia, and off-road vehicle fleets on different terrains with various traction coefficients. Significant interest has emerged in controlling the collective behavior of the ensemble systems, given the extensive number of applications. Due to the massive scale, control of ensembles is intractable at the individual level, being only practical at the population level, as shown in Figure \ref{fig:EnsembleBroad}.    
Specifically, measuring the states and correspondingly designing the controller of each individual in the ensemble is intractable/ infeasible in most cases. For example, a huge number of drones might only be controlled at the population level, as if it were an image of the entire swarm, given the availability of unique aggregated measurements and the limitation in sensing, perception, and computing resources. Although many researchers have investigated fundamental properties such as ensemble controllability, reachability, observability, and synchronizability in isolated or networked ensemble systems, these efforts have been mainly unconstrained. Moreover, the majority of the control problems in real deployments involve constrained scenarios, i.e., lower and upper bounds in command constraints, and obstacle avoidance for robots in state space. Therefore, to the best of our knowledge, this project proposes the first effort to address population-level ensemble control under constraints.

% \red{bullet points of contributions}
% \red{can you re-phase this following the abstract? }
The main contributions of this work are listed as follows:
\begin{itemize}      
    %\item Development of a set of constraints that portrays a polyhedron structure, converted from state space to moment space representation.  
    %\item Transformation in moment space of a set of polyhedral disjunctive constraints formulated with Big-M techniques, for an obstacle avoidance approach in ensemble control.
    \item  Development of a moment kernel transformation that maps parameterized ensemble constrained dynamics to moment space representation, including formulations for polyhedral way-point visiting and obstacle avoidance.      
    \item Representation of the expressive Signal Temporal Logic (STL) in the moment space, formulated to encode and enforce complex task specifications via a single shared controller synthesized from the constrained ensemble control framework.
    \item Implementation of the proposed approach in simulation and hardware experiments to demonstrate safe and efficient ensemble control of robots in constrained environments.  
\end{itemize}
This paper is organized as follows: In section \ref{sec:Related}, some ensemble control-related works are addressed. Sections \ref{sec:formulation} and \ref{sec:ensemble} present the problem formulation and moment kernel transformation, respectively, instantiated with the unicycle dynamics model. In section \ref{sec:transformation}, the moment transformation is explained for different constrained scenarios and obstacle avoidance. Later in section \ref{sec:STLmoment}, the approach is applied for STL specifications. Sections \ref{sec:simulation} and \ref{sec:hardwareExp} present the simulations and hardware experiments, respectively, followed by some concluding remarks in section \ref{sec:conclusions}.

%\red{an illustration of broadcasting control to an ensemble of heterogeneous robots}

\section{Related works} \label{sec:Related} 
% \red{andres: related work of distributional/ adaptive/ robust control in robotics}
%\red{can squeeze this if necessary}
Many research efforts have focused on studying basic properties of ensemble systems, such as controllability, reachability, and observability \cite{li2009ensemble,schonlein2021ensemble}. 
%It is worth mentioning that control of dynamical systems, particularly in the deployment of multiple robots, has been widely studied from the distributional, adaptive, and robust control. Considering the distributional control context, the work in \cite{boskos2023data} develops a distributionally robust coverage control algorithm based on data-driven and optimality guarantees for unknown spatial density, applied in the deployment of a team of robots.
Distributional, adaptive, and robust control have been widely used for multi-robots deployment. For these purposes, some studies are based on data-driven and optimality guarantees \cite{boskos2023data}, and Control Lyapunov Function (CLF) with Model Predictive Control (MPC) \cite{minniti2021adaptive}. Other approaches address uncertainties in a multi-robot network structure \cite{seraj2021adaptive} and combine robust control with the notion of ensemble \cite{becker2012approximate}.     
%Adaptive control has been used in \cite{minniti2021adaptive}, combining Control Lyapunov Function (CLF) for stability criteria with Model Predictive Control (MPC) to tackle the uncertainties presented in a quadrupedal robot. A more comprehensive adaptive control theory approach is shown by the authors in \cite{seraj2021adaptive}, developing a centralized and coordinated-control framework to address uncertainties in the network structure, which are translated into communications constraints in multi-robot teams. Other works, such as \cite{pastor2020ensemble}, present a robust MPC formulated to compute robust optimal trajectories while respecting inputs and state constraints. Parametric uncertainty of an unknown dynamical system is learned, providing a trade-off between actuation saturation and conservatism when handling modeling errors. In \cite{becker2012approximate} an approximate steering algorithm is applied on a nonholonomic unicycle in the presence of model perturbations, using robust control and the notion of ensemble control.
Approaches from other areas, including probability theory and kernel methods, have been developed to study ensemble systems \cite{zeng2016moment}. Meanwhile, formulation and techniques from ensemble systems are employed to study partial differential equations \cite{alleaume2024ensembles}. Distributional ensemble control and its application in robotic pattern control are studied as well \cite{li2025distributional}. However, constrained ensemble control, compared to its unconstrained counterpart, is far less investigated. The authors in \cite{aschenbruck2022consistency} study the constrained ensemble control with shared, but not population-level, control inputs, making it inapplicable to massive-scale ensemble systems. Similar techniques are applied in constrained re-entry flight \cite{selim2023safe} and robotic control \cite{chen2016optimal}, and hence inherit the same shortcomings. Rigorously speaking, those approaches still focus on individual members of the ensemble system, instead of treating the population as a whole. As a result, they are essentially multi-agent control techniques. 
Constrained optimal transport \cite{ekren2018constrained} and stochastic geometry \cite{chiu2013stochastic} are fundamental research areas with various applications. This project provides a novel perspective to study constrained ensemble control and its applications in many areas, including robotics and autonomy.

\section{Problem Formulation} \label{sec:formulation}

Consider the continuous-time dynamics in autonomous systems as
\begin{equation}\label{eq:dynamics}
    \dot{x} = f({x}, {u}, \psi)
\end{equation}
where ${x}\in\setX\subseteq\R^n$ is the state vector, such as the position and orientation of the robot, ${u}\in\R^m$ is the control input provided to the robot, and $\psi\in\setParam\subseteq \R^r$ is the parameter vector that captures variation in the ensemble.
We consider two models that are useful approximations of the dynamics of a wide range of robots \cite{cai2024evora}. 
Applicable to both differential-drive and legged robots, the unicycle model is defined as:
\begin{equation}\label{eq:unicycle}
    \begin{bmatrix}
        \dot{p}_{x}\\
        \dot{p}_{y}\\
        \dot{\theta}
    \end{bmatrix} = 
    \begin{bmatrix}
        \edit{\psi_{1}} \cdot v \cdot \sin{(\theta)}\\
        \edit{\psi_{1}} \cdot v \cdot \cos{(\theta)}\\
        \edit{\psi_{2}} \cdot \omega
    \end{bmatrix}
\end{equation}
where ${x}=[p_{x}, p_{y}, \theta]\tr$ contains the X, Y positions and yaw, ${u}=[v, \omega]\tr$ contains the commanded linear and angular velocities, $\psi=[\psi_{1}, \psi_{2}]\tr$ contains the linear and angular traction values $0\leq \psi_{1}, \psi_{2} \leq1$. Intuitively, traction captures the ``slip'', or the ratio of the achieved and commanded velocities. 
% \red{motivating examples: terrain agnostic off road autonomy: different terrains with different traction para; and delivery drones: payload mass varies} 
This difference between commands is related to several factors, such as off-road conditions (uneven surfaces and steep, muddy, sandy, and snowy terrains), two-/four- wheel drive, and electrical and mechanical losses throughout the engine, transmission, and shaft in ground vehicles. These can also be represented in a simplified fashion with the bicycle model, which is applicable for Ackermann-steering robots and is defined as:
\begin{equation}\label{eq:bicycle}
    \begin{bmatrix}
        \dot{p}_{x}\\
        \dot{p}_{y}\\
        \dot{\theta}
    \end{bmatrix} = 
    \begin{bmatrix}
        \edit{\psi_{1}} \cdot v \cdot \cos{(\theta)}\\
        \edit{\psi_{1}} \cdot v \cdot \sin{(\theta)}\\
        \edit{\psi_{2}} \cdot v \cdot \tan (\delta)/L%
    \end{bmatrix}
\end{equation}
where $L$ is the wheelbase, ${u}=[v, \delta]\tr$ contains the commanded linear velocity and steering angle, and ${\psi}$ plays the same role as in the unicycle model. The reference point for the bicycle model in \eqref{eq:bicycle} is located at the center between the two rear wheels. 
% \red{show another example of 6-DOF drone dynamics, e.g., the moment of inertia tensor/ mass of the drone can vary, possibly come from different masses of payload for delivery drones}. 
In the context of aerial vehicles, the translational and rotational 6-DOF quadrotor dynamics \cite{huang2023datt} is presented as follows
% in \eqref{eq:quad_dynamics}:
% \red{let use the compact form to save space: equation 1 in https://arxiv.org/pdf/2310.09053}
\begin{subequations}\label{eq:quad_dynamics}
\begin{align}
&\dot{\mathbf{x}} = \mathbf{v}, \quad m(\psi) \dot{\mathbf{v}} = m(\psi) \mathbf{g} + R\,\mathbf{e}_z\, T, \\&\dot{R} = R\,S({\boldsymbol{\omega}}), \quad J(\psi) \dot{\boldsymbol{\omega}} = J(\psi) \boldsymbol{\omega}\times\boldsymbol{\omega} + \boldsymbol{\tau}. 
\end{align}
\end{subequations}
where $\mathbf{x},\,\mathbf{v},\,\mathbf{g} \in \mathbb{R}^3$ are the position, velocity, and gravity vectors in the world frame, respectively;
$R$ is the body-to-world rotation matrix and 
$\boldsymbol{\omega} \in \mathbb{R}^3$ is the angular velocity in the body frame. Operator $S(\cdot)$ maps $w$ from $\mathbb{R}^3$ to the skew-symmetric matrix.
The vector $\mathbf{e}_z = [0\;0\;1]^\top$ indicates the body $z$-axis. The total thrust and body torque are denoted by $T$ and $\boldsymbol{\tau}$, respectively.  For the variation of dynamics, the parameter $\psi$ can be applied to the mass $m(\psi)$ and inertia matrix $J(\psi)$, e.g., varying load for package-delivery drones.
% \red{do you know what $S(\omega)$ means in the reference?}

\begin{comment}
    \begin{equation}
\ddot{\mathbf{p}} = 
g\,\mathbf{e}_3 
+ \frac{1}{m}\,R(\boldsymbol{\eta})
\begin{bmatrix}
0 \\[3pt] 0 \\[3pt] -u_1
\end{bmatrix}
\label{eq:translational}
\end{equation}
\begin{equation}
\ddot{\boldsymbol{\eta}} =
\begin{bmatrix}
\dfrac{I_{yy}-I_{zz}}{I_{xx}}\,\dot{\theta}\dot{\zeta} \\[8pt]
\dfrac{I_{zz}-I_{xx}}{I_{yy}}\,\dot{\phi}\dot{\zeta} \\[8pt]
\dfrac{I_{xx}-I_{yy}}{I_{zz}}\,\dot{\phi}\dot{\theta}
\end{bmatrix}
+
\begin{bmatrix}
\dfrac{L}{I_{xx}} & 0 & 0 \\[6pt]
0 & \dfrac{L}{I_{yy}} & 0 \\[6pt]
0 & 0 & -\dfrac{1}{I_{zz}}
\end{bmatrix}
\begin{bmatrix}
u_2 \\[4pt] u_3 \\[4pt] u_4
\end{bmatrix},
\label{eq:rotational}
\end{equation}
where \(\mathbf{p} = [x\;y\;z]^\top\) represents the inertial position,
\(R(\boldsymbol{\eta})\) is the rotation matrix from the body to the inertial frame,
\(\boldsymbol{\eta} = [\phi\;\theta\;\zeta]^\top\) are the Euler angles and
\(u_1\) is the total thrust. Control torques are \(u_2, u_3, u_4\). For dynamics variation, the parameter $\psi$ can be applied to the mass $m(\psi)$ and moments of inertia \(I_{xx}(\psi), I_{yy}(\psi), I_{zz}(\psi)\), i.e., varying load for package-delivery drones.
\end{comment}

%============================
\subsection{Ensemble system formulation of robot populations}

To put our analysis in a more general setting, we note that the system in \eqref{eq:dynamics} is essentially a parameterized control system, which we refer to as an \textit{ensemble system}. With the system parameter $\psi$ varying on the space $\setParam$, the ensemble state defines a function $x_t(\cdot)\doteq x(t,\cdot):\setParam\rightarrow\setX$, which outputs the state $x_t(\psi)$ of the system in the ensemble with the dynamics characterized by $\psi$. Formally, the ensemble system is formulated as a control system evolving on a space $\mathcal{F}(\setParam,\setX)$ of $\setX$-valued functions defined on $\setParam$, denoted more specifically as $\frac{d}{dt}x(t,\psi) = f(x(t,\psi),u(t),\psi)$. A major advantage of this functional formulation is its ability to model arbitrarily large ensembles of control systems. Indeed, when $\setParam$ is an infinite set, this ensemble consists of infinitely many heterogeneous control systems. 

% we consider a control system
% \begin{equation}\label{eq:ensemble}
%     \frac{d}{dt}x(t,\psi) = f(x(t,\psi), \psi, u(t))
% \end{equation} 
% parameterized by a random variable $\psi$ drawn from a probability distribution $\mu$ on $\Omega\subset\mathbb{R}$, where $x(t,\psi)\in\mathbb{R}^n$ is the system state, $u(t)\in\mathbb{R}^r$ is the control input, and $f(\cdot,\psi,u(t))$ is a vector field on $\mathbb{R}^n$ for each $\psi\in\mathbb{R}^d$ and $u(t)\in\mathbb{R}^r$. Note that our approach works for cases with $\psi$ as a scalar, vector, or tensor. We refer to the parameterized system in \eqref{eq:ensemble} as an \textit{ensemble system}, and both the unicycle model in \eqref{eq:unicycle} and the Ackermann-steering robot model in \eqref{eq:bicycle} are in the form of ensemble systems. 

In practice, safety constraints are generally expected for control systems. Such a constraint is usually represented in terms of the sublevel set $\{x\in\setX: h(x)\leq0\}$ of a function $h:\setX\rightarrow\mathbb{R}$. In the ensemble system case, this constraint is required to be enforced for every individual system so that $h(x_t(\psi))\leq0$ for all $\psi\in\setParam$. This poses a significant challenge to the control of the ensemble system. In this work, we will propose a novel safe control approach inclusive for arbitrarily large heterogeneous ensemble systems. %The central idea is to transform the system, together with the safety constraints, to another domain to mitigate the challenge caused by the variation in $\psi$. 

%To pave the way to the development of this transform, we first carry out a functional interpretation of the ensemble system in \eqref{eq:ensemble}. As $\psi$ varies on $\Omega$, the ensemble state becomes an $\mathbb{R}^n$-valued function defined on $\mathbb{R}^d$, which we denote by $x_t(\cdot)\doteq x(t,\cdot):\mathbb{R}^d\rightarrow\mathbb{R}^n$. To take the distribution of $\psi$ into this functional formulation, we impose the condition $x_t\in L^2(\Omega,\mu)$, the space of square-integrable $\mathbb{R}^n$-valued functions define on $\mathbb{R}$ with respect to the measure $\mu$, meaning $\|x_t\|^2=\int_{\Omega}|x_t|^2d\mu<\infty$, where $|\cdot|$ denotes a norm on $\mathbb{R}^n$. Interpreted from this perspective, the ensemble system in \eqref{eq:ensemble} is a control system defined on the infinite-dimensional function space $L^2(\Omega,\mu)$. Translated into this setup, the aforementioned challenge is indeed the manifestation of the infinite dimensionality of the ensemble system. 

%This functional interpretation inspires the introduction of the notion of moments for the study of autonomy safety for the ensemble system in \eqref{eq:ensemble}.

%%%%%%%%%%%%%%%%%%%%%%%%%%%%%%%%
\section{Moment kernel reduction of ensemble systems} \label{sec:ensemble}

In this section, we will propose a new kernelization paradigm to construct a reduced model for the ensemble system. The development is enabled by the functional interpretation of the ensemble system. The main idea centers around representing the ensemble state function in terms of a sequence, which then allows the low dimensional truncation approximation of the ensemble system. 

%\red{lets try to squeeze this section to about 1 page}

%===============================
\subsection{Moment kernel transform of ensemble systems} 
%under safety constraints}
To illuminate the main idea in the simplest setting, 
%To construct a reduced representation of the ensemble system, we leverage its functional interpretation as a control system defined on $L^2(\Omega,\mu)$. To illuminate the main idea, 
we start from the case where $x_t$ is in the space $L^2(\Omega)$ of real-valued square-integrable functions defined on $\Omega$. Since $L^2(\Omega)$ is a separable Hilbert space \cite{Folland1999}, there exists an orthonormal basis $\{\phi_k\}_{k\in\mathbb{N}}$, through which we introduce the notion of \textit{moments} for the ensemble system in \eqref{eq:dynamics} as
\begin{align}
    m_k(t)=\langle \phi_k,x_t\rangle=\int_\Omega\phi_k(\psi)x_t(\psi)d\psi
    \label{eq:moment}
\end{align}
for all $k\in\mathbb{N}$. The \textit{moment sequence} $m(t)=\big(m_k(t)\big)_{k\in\mathbb{N}}$ is essentially the Fourier coefficients of the function $x_t$. Therefore, $m(t)$ is in one-to-one correspondence to $x_t$ and satisfies $\|m(t)\|^2=\sum_{k\in\mathbb{N}}|m_k(t)|^2=\int_\Omega x_t^2(\beta)d\beta=\|x_t\|^2<\infty$ \cite{Folland1999}. Formally, this shows that the space of moment sequences coincides with $\ell^2$, the space of square-summable sequences, and the \textit{moment transform} $\mathcal{K}:x_t\mapsto m(t)$ is an isometric isomorphism. In the general case where $x_t$ is $\mathbb{R}^n$-valued, the notion of moments in \eqref{eq:moment} can be defined in a component-wise manner so that the moment sequence $m(t)$ becomes an $\mathbb{R}^n$-valued $\ell^2$-sequence and all the aforementioned properties remains satisfied. 

More importantly, $\ell^2$ is a reproducing kernel Hilbert space (RKHS) with the reproducing kernel given by the $\ell^2$-inner product \cite{Paulsen2016}. From this perspective, $m(t)$ gives rise to a kernel representation of the ensemble state $x_t$. To benefit ensemble control from this kernelization, we will derive the moment kernelized ensemble system. 

% Because $\{\phi_k\}_{k\in\mathbb{N}}$ is a basis for $L^2(\Omega,\mu)$, the ensemble state $x_t$ is in one-to-one correspondence to the \textit{moment sequence} $m(t)=\big(m_k(t)\big)_{k\in\mathbb{N}}$. In addition, by Parseval's identity, its orthonormal property implies $x_t=\sum_{k=0}^\infty m_k(t)\phi_k$ in the $L^2$ sense, meaning $\|x_t-\sum_{k=0}^\infty m_k(t)\phi_k\|^2=\int_\Omega|x_t-\sum_{k=0}^\infty m_k(t)\phi_k|^2d\mu=0$, and $\|m(t)\|^2=\sum_{k=0}^\infty |m_k(t)|^2<\infty$. The square summability of $m(t)$ further indicates that moment sequences are in the $\ell^2$-space, which is a reproducing kernel Hilbert space (RKHS). Indeed, the \textit{moment kernel transform} $x_t\mapsto m(t)$ gives rise to a kernel representation of the ensemble system in \eqref{eq:ensemble}. In the general case where $n>1$, the definition of moments in \eqref{eq:moment} applies to each component of $x(t,\beta)=\big(x_1(t,\beta),\dots,x_n(t,\beta)\big)$, which results in an $\mathbb{R}^n$-valued moment sequence $m(t)=\big(m_k(t)\big)_{k\in\mathbb{N}}$.

%=================================
\subsection{Moment kernelized unicycle ensembles}

We will showcase the derivation of the moment system using the unicycle ensemble in \eqref{eq:bicycle}, with $\psi_1=\psi_2$ rescaled to $\beta\in[-1,1]$. To facilitate efficient moment calculation, we first apply the change of coordinates $(p_x,p_y,\theta)\mapsto(p_x,p_y,\cos\theta,\sin\theta)\doteq z$. Under the $z$-coordinates, the unicycle dynamics is represented by a bilinear system 
\begin{align}
    \frac{d}{dt}z(t,\beta)&=\beta \Big(v\left[\begin{array}{cc} 0 & \Lambda \\ 0 & 0 \\ \end{array}\right]+\omega\left[\begin{array}{cc} 0 & 0 \\ 0 & J \\ \end{array}\right] \Big)z(t,\beta)\nonumber\\
    %=v\left[\begin{array}{cccc} 0 & 0 & 0 & \beta \\ 0 & 0 & \beta & 0 \\  0 & 0 & 0  & 0 \\ 0 & 0 & 0 & 0\end{array}\right]z+\omega \left[\begin{array}{cccc} 0 & 0 & 0 & 0 \\ 0 & 0 & 0 & 0 \\  0 & 0 & 0 & -\beta \\ 0 & 0 & \beta & 0\end{array}\right]z\\
    &\doteq\beta \big(v B_1+ \omega B_2\big)z(t,\beta) \label{eq:unicycle_bilinear}
\end{align}
where $\Lambda=\left[\begin{array}{cc} 0 & 1 \\ 1 & 0 \\ \end{array}\right]$, $J=\left[\begin{array}{cc} 0 & -1 \\ 1 & 0 \\ \end{array}\right]$, and $0\in\mathbb{R}^{2\times2}$ is the zero matrix. We then choose $\{\phi_k\}_{k\in\mathbb{N}}$ to be the set of Legendre polynomials. Notably, Legendre polynomials satisfy Bonnet's recursive formula 
\begin{equation} \label{eq:RecurrenceFinal}
\beta\phi_k(\beta) = a_k\phi_{k+1}(\beta) + c_k\phi_{k-1}(\beta),
\end{equation}
where $c_0=0$, $c_k=a_{k-1}=k/\sqrt{4k^2-1}$ for all $k\geq1$, and we define $\phi_{-1}=0$ and $\phi_0=1$. Integrating this recursive relation into the moment system derivation yields for all $k\in\mathbb{N}$
\begin{align*}
    &\frac{d}{dt}m_k(t)=\frac{d}{dt}\int_{\Omega}\phi_k(\beta)x_t(\beta)d\beta=\int_\Omega\phi_k(\psi)\frac{d}{dt}x_t(\beta)d\beta\\
    &=\int_\Omega\beta\phi_k(\beta)\big(v B_1+\omega B_2\big)z(t,\beta)d\beta\\
    &=\int_\Omega\big(a_k\phi_{k+1}(\beta) + c_k\phi_{k-1}(\beta)\big)\big(v B_1+\omega B_2\big)z(t,\beta)d\beta\\
    &=a_kvB_1\int_\Omega\phi_{k+1}(\psi)z(t,\beta)d\beta+c_k\omega B_2\int_\Omega\phi_{k-1}(\beta)z(t,\beta)d\beta\\
    &=a_kvB_1m_{k+1}(t)+c_k\omega B_2m_{k-1}(t),
\end{align*}
 which remains a bilinear system. In the matrix (tensorial) form, the moment kernelized unicycle ensemble is given by
\begin{align}
    \frac{d}{dt}m(t)&=R\otimes\big(vB_1+\omega B_2\big)(a\circ m(t)\big)\nonumber\\
    &\qquad\quad\quad+L\otimes\big(vB_1+\omega B_2\big)(c\circ m(t)\big)\nonumber\\
    &\doteq v\mathcal{B}_1m(t)+\omega \mathcal{B}_2m(t),\label{eq:moment_system}
\end{align}
where $L:(m_0(t),m_1(t),m_2(t),\dots)\mapsto(m_1(t),m_2(t),m_3(t),\dots)$ and $R:(m_0(t),m_1(t),m_2(t),\dots)\mapsto(0,m_1(t),m_2(t),\dots)$ are the left- and right-shift operators on $\ell^2$, respectively, $\otimes$ and $\circ$ denote the Kronecker and Hadamard products of matrices, respectively, and $a=(a_k)_{k\in\mathbb{N}}$ and $c=(c_{k+1})_{k\in\mathbb{N}}$ are the sequences generated by the Bonnet's recursive relation in \eqref{eq:RecurrenceFinal}.

\section{Constrained Ensemble Control: Transforming constraints from state space to moment space} \label{sec:transformation}

Starting from this section, we integrate state constraints into ensemble robot control. The essential step is to represent the constraints in terms of moment sequences. In most practical cases, state constraints constitute smooth submanifolds of the system state spaces, determined by the regular level sets or sublevel sets of smooth functions. It is a well-known result in differential geometry that every smooth manifold admits a triangulization \cite{Munkres1967}. Therefore, it suffices to consider polyhedral region constraints.

% \red{we should focus more on this section than the unconstrained control section}
\subsection{Polyhedral region exploration}
A polyhedral region in $\mathbb{R}^n$ is the intersection of the sub- and/or super-level sets of a family of linear functionals on $\mathbb{R}^n$, say $S_j=\{x\in\mathbb{R}^n:b_j\leq\sum_{i=1}^na_{ij}x_j\leq c_j\}$ for $b_j,c_j\in\mathbb{R}$ and $j=1,\dots,r$. It can then be represented in the matrix form as $S=\{x\in\mathbb{R}^n:b\leq Ax\leq c\}=\bigcap_{i=1}^rS_j$, where $A\in\mathbb{R}^{r\times n}$ has the $(i,j)$-th entry $a_{ij}$, $b$ and $c$ in $\mathbb{R}^r$ have the $i^{\rm th}$ entries $b_i$ and $c_i$, respectively, and $\leq$ evaluated in the component-wise manner. 

\subsubsection{Moment kernelization of polyhedra}
We consider the case where each robot in the ensemble in \eqref{eq:bicycle} is required to move through the region $S$. To derive the moment kernel representation of $S$, it suffices to work on each $S_j$. We define the positive and negative parts of the function $\phi_k$ to be $\phi_k^+=\max\{\phi_k,0\}$ and $\phi_k^-=\max\{-\phi_k,0\}$, respectively, for each $k\in\mathbb{N}$. Then, we have $\phi_k=\phi_k^+-\phi_k^-$ so that $m_k(t)$ can be represented as 
$$m_k(t)=\int_\Omega\phi_k^+x_td\mu-\int_\Omega\phi_k^-x_td\mu.$$ 
The state constraint $b\leq Ax_t\leq c$ yields $\phi_k^{\pm}b\leq\phi_k^{\pm} Ax_t\leq \phi_k^{\pm}c$ such that $\int_{\Omega}\phi_k^{\pm}bd\mu \leq\int_\Omega\phi_k^{\pm} Ax_td\mu\leq\int_\Omega \phi_k^{\pm}cd\mu$. This leads to the constraint on $m_k(t)$, given by
\begin{align*}
    b\int_{\Omega}\phi_k^+d\mu -c\int_{\Omega}\phi_k^-d\mu&\leq Am_k(t) \\
    &\qquad\leq c\int_{\Omega}\phi_k^+d\mu -b\int_{\Omega}\phi_k^-d\mu
\end{align*}
Note that $\phi_k^{\pm}$ are essentially the restriction of $\pm \phi_k$ on the sets $D_k^{\pm}$ where they take non-negative values. In this case, $D_k^{\pm} \subseteq \Omega$ and are defined by the roots of the basis $ \phi_k$, i.e, for Legendre polynomials $\Omega \in [-1,1]$. Therefore, we have $\int_\Omega\phi_k^{\pm}d\mu=\pm\int_{D_{k}^{\pm}}\phi_kd\mu\doteq m_k^{\pm}$, which can be easily evaluated numerically in practice. Let $m^{\pm}=(m_k^{\pm})_{k\in\mathbb{N}}$ be the ``positive/negative moment sequence'', then the moment kernel representation of the polyhedral region $S$ follows
\begin{align}
    (I\otimes b)(m^+&\otimes\mathbf{1})-(I\otimes c)(m^-\otimes\mathbf{1})\leq(I\otimes A)m(t)\nonumber\\
    &\leq(I\otimes c)(m^+\otimes\mathbf{1})-(I\otimes b)(m^-\otimes\mathbf{1}), \label{eq:polyhedron_moment}
\end{align}
where $\mathbf{1}\in\mathbb{R}^r$ is the all-one vector. 

% \red{after all, this is our main contribution; if necessary, can give an example/ visualization of 2nd order Legendre polynomial}
\begin{example}

A polyhedron is represented by $b \leq Ax \leq c$, i.e., a set of constraints in state space,
\begin{comment}
\begin{equation} \label{eq:SetPoly}
    \begin{split}
        &-1<2x_1+1x_2<13\\
        &-2<x_2<2
    \end{split}
\end{equation}
\end{comment}
where $x \in \mathbb{R}^2$, $A=[2 \hspace{0.5em} 1;0 \hspace{0.5em} 1]$, $b=[-1, -2]^{\top}$ and $c=[13, 2]^{\top}$. 
The moment transformation considers $\phi_k^+(\mu)$ and $\phi_k^-(\mu)$ applied in the previous inequalities. For the first constraint, the transformation is as follows:
\begin{align*}
    &c_1\phi_k^-(\mu)  < \phi_k^-(\mu)(A_{1,1}x_1+A_{1,2}x_2) < b_1\phi_k^-(\mu),\\
    &b_1\phi_k^+(\mu)  < \phi_k^+(\mu)(A_{1,1}x_1+A_{1,2}x_2) < c_1\phi_k^+(\mu).
\end{align*}
Adding them and applying integration yields
\begin{comment}
    \begin{align*}            
    &c_{1}\int_{\mu_1}^{\mu_2} \phi_2^-(\mu) d\mu +b_1\left[\ \int_{\mu_l}^{\mu_1} \phi_2^+(\mu)d\mu + \int_{\mu_2}^{\mu_u} \phi_2^+(\mu)d\mu \right]\\
    &<A_{1,1}m_{k1} + A_{1,2}m_{k2} \\ 
    & < b_1\int_{\mu_1}^{\mu_2} \phi_2^-(\mu) d\mu +c_1\left[\ \int_{\mu_l}^{\mu_1} \phi_2^+(\mu)d\mu + \int_{\mu_2}^{\mu_u} \phi_2^+(\mu)d\mu \right], 
\end{align*}
\end{comment}
\begin{align*}
    &\int_{\mu_l}^{\mu_u} \left(\ c_1\phi_k^-(\mu) +b_1\phi_k^+(\mu) \right)d\mu \le \\ 
    &A_{1,1} \underbrace{\int_{\mu_l}^{\mu_u} \left(\    x_1 (\phi_k^+(\mu) + \phi_k^-(\mu)\right)d\mu}_{m_{k1}}  \\ 
    & +A_{1,2} \underbrace{\int_{\mu_l}^{\mu_u} \left(\ x_2 (\phi_k^+(\mu) + \phi_k^-(\mu) \right) d\mu}_{m_{k2}}  \\
    &\le \int_{\mu_l}^{\mu_u} \left(\ c_1\phi_k^+(\mu) +b_1\phi_k^-(\mu) \right)d\mu.
\end{align*}
With $k=2$, to get the $2^{\text{nd}}$ order moment representation, the transformation is based on the second-order Legendre polynomial, which is depicted in Figure \ref{fig:LegPolySecond}. Higher order ones are shown in Figures \ref{fig:LegPolyThird} and \ref{fig:LegPolyForth}, for the $3^{\text{rd}}$ and $4^{\text{th}}$ order polynomial, respectively.  
\begin{figure}
    \centering
    \includegraphics[width=1\linewidth]{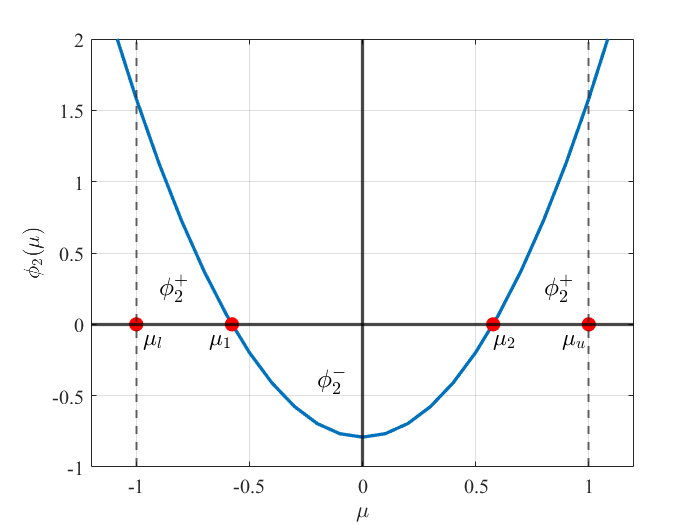}
    \caption{
    The second-order Legendre polynomial is given by 
    $\phi_2(\mu) = \tfrac{1}{2}(3\mu^2 - 1)$, with roots at $\mu_1 = -\tfrac{1}{\sqrt{3}}$ and $\mu_2 = \tfrac{1}{\sqrt{3}}$, defined over $-1=\mu_l \leq \mu \leq \mu_u=1$.
    % \red{is it possible that we can annotate the zero points here, e.g, 0.57}
    }
    \label{fig:LegPolySecond}
\end{figure}
\begin{figure}[t]
    \centering
    % Each subfigure takes ~47% of column width to fit nicely side by side
    \begin{subfigure}{0.49\linewidth}
        \centering
        \includegraphics[width=\linewidth]{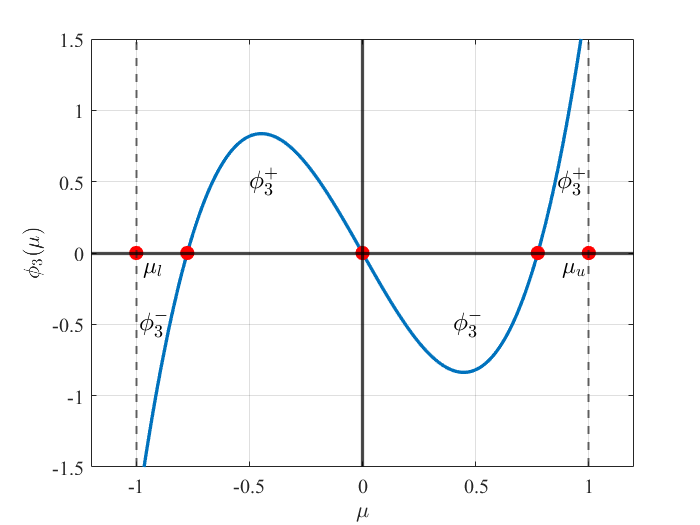}
        \caption{Third-order polynomial $\phi_{3}(\mu) = \frac{1}{2}\left(5\mu^{3} - 3\mu\right)$ with roots at $\mu = 0$ and $\mu = \pm0.77.$}
        \label{fig:LegPolyThird}
    \end{subfigure}
    \hfill
    \begin{subfigure}{0.49\linewidth}
        \centering
        \includegraphics[width=\linewidth]{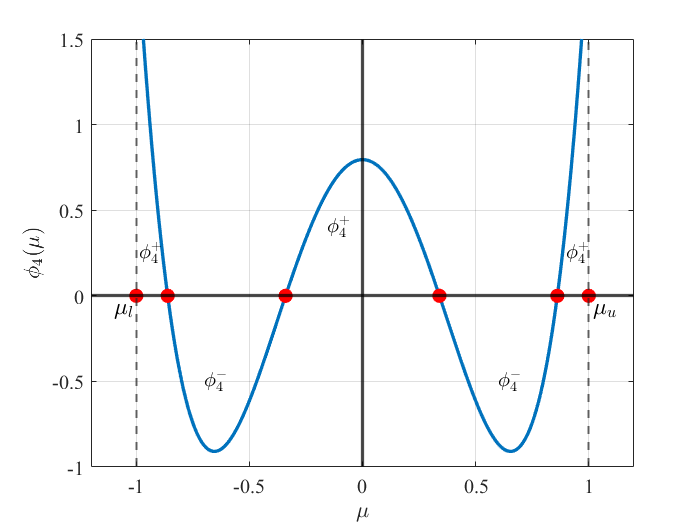}
        \caption{Forth-order polynomial $\phi_{4}(\mu) = \frac{1}{8}\left(35\mu^{4} - 30\mu^{2} + 3\right)$ with roots at $\mu = \pm0.86$ and $\mu = \pm0.34.$}
        \label{fig:LegPolyForth}
    \end{subfigure}
    \caption{Higher order Legendre polynomials, defined over $-1=\mu_l \leq \mu \leq \mu_u=1$.}
    \label{fig:comparisonBox}
\end{figure}
The positive and the negative parts of $\phi_2(\mu)$ are denoted as $\phi_2^+$ and $\phi_2^-$, respectively, and are correspondingly evaluated in the integrals, as presented below:   
\begin{align*}            
    &c_{1}\int_{\mu_1}^{\mu_2} \phi_2^-(\mu) d\mu +b_1\left[\ \int_{\mu_l}^{\mu_1} \phi_2^+(\mu)d\mu + \int_{\mu_2}^{\mu_u} \phi_2^+(\mu)d\mu \right]\\
    &<A_{1,1}m_{k1} + A_{1,2}m_{k2} \\ 
    & < b_1\int_{\mu_1}^{\mu_2} \phi_2^-(\mu) d\mu +c_1\left[\ \int_{\mu_l}^{\mu_1} \phi_2^+(\mu)d\mu + \int_{\mu_2}^{\mu_u} \phi_2^+(\mu)d\mu \right], 
\end{align*}
where $m_{k1}$ and $m_{k2}$ are the $k^{\text{th}}$ order moment variables for $x_1$ and $x_2$, respectively. The process is repeated for the second state space constraint. Finally, the polyhedron constraints in moment space are: $\mathbf{b}_m \;\le\; A\mathbf{m} \leq \mathbf{c}_m $, where $\mathbf{b}_m = [\,-8.5,\,-2.4 \,]^{\top}, \mathbf{m} = [\,m_{k1},\, m_{k2}\,]^{\top},$ and $
\mathbf{c}_m = [\,8.5,\, 2.4\,]^{\top}$.
\begin{comment}
\begin{align*}
    &2\int_{-1}^{1} \phi_k^-(\mu) d\mu  -2\int_{-1}^{1} \phi_k^+(\mu) d\mu < 0m_{k_1} + 1m_{k_2} < \\ 
    &-2\int_{-1}^{1} \phi_k^-(\mu) d\mu  +2\int_{-1}^{1} \phi_k^+(\mu) d\mu
\end{align*}
Deploying until the second order Legendre polynomial, i.e., $\phi_0(\mu)=1$, $\phi_1(\mu)=\mu$ and $\phi_2(\mu)=\frac{1}{2} \left(3\mu^2-1\right)$; the polyhedron constraints in moment space are shown below:
\begin{align*}
&\ell_k \;\le\;
\begin{bmatrix}
2 & 1 \\
0 & 1
\end{bmatrix}
\begin{bmatrix}
m_{k1} \\
m_{k2}
\end{bmatrix}
\;\le\;
u_k,
\hspace{0.5em}k = 0,1,2,\\
&\ell_0 =
\begin{bmatrix}
-1.2142 \\
-2.8284
\end{bmatrix},\hspace{0.5em}
u_0 =
\begin{bmatrix}
10.3848 \\
2.8284
\end{bmatrix},\\
&\ell_1 =
\begin{bmatrix}
-8.5732 \\
-2.4495
\end{bmatrix},\hspace{0.5em}
u_1 =
\begin{bmatrix}
8.5732 \\
2.4495
\end{bmatrix},\\
&\ell_2 =
\begin{bmatrix}
-8.5201 \\
-2.4344
\end{bmatrix},\hspace{0.5em}
u_2 =
\begin{bmatrix}
8.5201 \\
2.4343
\end{bmatrix}.
\end{align*}
\end{comment}
\end{example}
\subsubsection{Constrained control of unicycle ensembles}
In the (bilinearized) unicycle ensemble in \eqref{eq:unicycle_bilinear}, the first two components of the system state $z(t,\cdot)$ characterize the workspace locations of the unicycles. Therefore, the constraint $S\subset\mathbb{R}^4$ is necessarily a polyhedron embedded into $\mathbb{R}^2$. Correspondingly, the matrix $A\in\mathbb{R}^{r\times 4}$ in $S$ is in the block form of $A=[\, \tilde A\mid 0\, ]$ for some $\tilde A\in\mathbb{R}^{r\times 2}$. 

Let $\tilde m(t)=\big(\tilde m_k(t)\big)_{k\in\mathbb{N}}$ be the $\mathbb{R}^2$-valued sequence consisting of the first two components $\tilde m_k(t)$ of $m_k(t)$ and $\tilde m_F$ the moment sequence of the desired workspace location for the unicycle ensemble. Then, the polyhedral region exploration task can be formulated as the optimal control problem for the moment system as follows
\begin{align}
   \min_{u,v:[0,T]\rightarrow\mathbb{R}}&\ \|\tilde m(T)-\tilde m_F\|^2 \nonumber\\
   & \hspace{-4.5em} {\rm s.t.} \hspace{0.5em} \frac{d}{dt}m(t)=u(t)\mathcal{B}_1m(t)+u(t)\mathcal{B}_2m(t), \nonumber\\
    & \hspace{-4.5em} (I\otimes b)(m^+\otimes\mathbf{1})-(I\otimes c)(m^-\otimes\mathbf{1})  \label{eq:poly_explor}
    \leq \\ & \hspace{-4.5em} (I\otimes \tilde A)\tilde m(t) \leq (I\otimes c)(m^+\otimes\mathbf{1})\nonumber
    -(I\otimes b)(m^-\otimes\mathbf{1}). \nonumber
\end{align}

\subsection{Obstacle avoidance} \label{ObsAvoid}

Now, we reverse the engineering to require the ensemble system to avoid polyhedral obstacles, formulated as $O = \{x\in\mathbb{R}^n: Ax \ge b\}$, where $A=[a_1 \hspace{0.3em} a_2 \hspace{0.3em} \cdots \hspace{0.3em} a_d]^{\top}$ with $d$ denoting the number of facets of the polyhedron obstacle. Then the safe zone, the complement set of the obstacle, is formulated by a set of \textit{disjunctive constraints} parameterized as $\bigcup_{i=1}^dS_i$, with $S_i=\{x\in\mathbb{R}^n:a_i^\top x\leq b_i\}$.
% and $d$ denotes the number of facets of the polyhedron obstacle. 
% In matrix form, the polyhedron obstacle is formulated as $\{x\in\mathbb{R}^n: Ax \ge b\}$, where $A=[a_1 \hspace{0.3em} a_2 \hspace{0.3em} \cdots \hspace{0.3em} a_d]^{\top}$}. \red{suggesting the following way: 
% \red{following my suggestion, please; the obstacle is $Ax\ge b$, the safe zone, is its complement set, formulated as a set of disjunctive constraints, is $a_i^\top x\leq b_i$}. With the polyhedron obstacles formulated as $O = \{x|Ax\ge b\}$, the safe zone, parametrized as $S = \vee_iS_i$, with $S_i=\{x\in\mathbb{R}^n:a_i^\top x\leq b_i\}$ ; then adding how $a_i$ and $A$ are related.}
%$i$ the index for the edge $i \in d$
%$j$ the index for the variable $x_j$, being $x \in \mathbb{R}^n$. 
%$S_i=\{x\in\mathbb{R}^n:A_ix\leq b_i\}$ for $i=1,\dots,d$. 
We then apply the big-M method to tackle this obstacle avoidance problem, which reformulates the disjunctive constraints to 
\begin{equation}\label{eq:setPolyObsArr}
-M \leq a_ix(t,\beta)+ Mz_i \leq b_i + M 
\end{equation}
for $i=1,\dots,d$, where $M>>0$, and $z_i\in\{0,1\}$ such that 
$$\sum_{i=1}^dz_i \ge 1.$$ 
Note that $z_i=1$ means that $a_i^\top x\leq b_i$ is active  \cite{garcia2023combinatorial}. 
The transformed constraints in \eqref{eq:setPolyObsArr} are in the polyhedral type considered in the previous section, and hence their moment kernel representations are in the form of \eqref{eq:polyhedron_moment} as 
\begin{align}
&-(I\otimes M)(m^+\otimes \mathbf{1})-(I\otimes(b_i+M))(m^-\otimes\mathbf{1}) \nonumber\\
&\leq(I\otimes a_i)m(t) z_i \\
&\leq(I\otimes(b_i+M))(m^+\otimes\mathbf{1})+(I\otimes M)(m^+\otimes \mathbf{1}). \nonumber
\end{align}

\begin{comment}
  $$-(I\otimes M)(m^+\otimes \mathbf{1})-(I\otimes(b_i+M))(m^-\otimes\mathbf{1})
\leq(I\otimes a_i)m(t) z_i
\leq(I\otimes(b_i+M))(m^+\otimes\mathbf{1})+(I\otimes M)(m^+\otimes \mathbf{1}).$$  
\end{comment}

% \red{we probably should do an example for this equation as well}

\begin{example}
The following constraints $\{x\in\mathbb{R}^2: Ax \ge b\}$ represent a square-shaped obstacle: $A=[1 \hspace{0.5em} 0;0 \hspace{0.3em} -1;-1 \hspace{0.5em} 0;0 \hspace{0.5em} 1]$ and $b=[x_{1}^L, -x_{1}^U,x_{2}^L,-x_{2}^U]^{\top}$, where $x_{1}^L \le x_1 \leq x_{1}^U$ and $x_{2}^L \le x_2 \leq x_{2}^U$. In this example, $2 \le x_1 \leq 6$ and $1 \le x_2 \leq 3$. The safe set, as the complementary set of the obstacle, is formulated as four disjunctive constraints, shown below according to the aforementioned reformulation: 
\begin{align*}
&-M \leq A_{1,1}x_1 +A_{1,2}x_2 +Mz_1 \leq x_{1}^L+M, \\
& -M \leq A_{2,1}x_1 +A_{2,2}x_2+Mz_2 \leq -x_{2}^U+M, \\ 
& -M \leq A_{3,1}x_1 +A_{3,2}x_2+Mz_3 \leq -x_{1}^U+M, \\
& -M \leq A_{4,1}x_1 +A_{4,2}x_2 + Mz_4 \leq x_{2}^L + M. \\    
\end{align*}
\begin{comment}
$-M \leq A_{1,1}x_1 +A_{1,2}x_2 +Mz_1 \leq x_{1}^L+M$, OR $-M \leq A_{2,1}x_1 +A_{2,2}x_2+Mz_2 \leq -x_{2}^U+M$, OR $-M \leq A_{3,1}x_1 +A_{3,2}x_2+Mz_3 \leq -x_{1}^U+M$, OR $-M \leq A_{4,1}x_1 +A_{4,2}x_2 + Mz_4 \leq x_{2}^L + M$. 
\end{comment}
Considering the zero order Legendre polynomial, $\phi_0(\mu)=1$, the moment transformation for the first disjunctive constraint is shown as follows. For simplicity, the $\mu$ dependence has been temporarily omitted during the derivation: 
\begin{align*}
&-M \int_{\mu_l}^{\mu_u} \phi_0^+d\mu +
\xi\int_{\mu_l}^{\mu_u} \phi_0^-d\mu \leq A_{1,1}m_{k1} +A_{1,2}m_{k2} \\ &+M \left[ \int_{\mu_l}^{\mu_u} \phi_0^+d\mu  +   \int_{\mu_l}^{\mu_u} \phi_0^-d\mu \right]     z_1 \leq \\ &\xi\int_{\mu_l}^{\mu_u} \phi_0^+d\mu  -M \int_{\mu_l}^{\mu_u} \phi_0^-d\mu, 
\end{align*}
%\red{does not read right, on the left, original equation was $-M$, now it is $x_1^L$} 
where $\xi=
x_{1}^L+M$ and $m_{k1}$ and $m_{k2}$ are the zero order moment variables for $x_1$ and $x_2$, respectively; and $z_1$ is the binary variable associated with the edge. The rest of the constraints undergo the same procedure, which yields the following set of disjunctive constraints in moment space representation: $(-28.3)\,\mathbf{1}_4 \;\le\; A\mathbf{m} + 28.3\,\mathbf{z} \;\le\; \mathbf{b}$, where $\mathbf{m} = [\,m_{k1},\, m_{k2}\,]^{\top},
\mathbf{z} \in \{0,1\}^4, 
\mathbf{b} = [\,31.1,\, 24.1,\, 19.8,\, 29.7\,]^{\top}$ and $\mathbf{1}_4 = [\,1,\, 1,\, 1,\, 1\,]^{\top}$. %\red{this is wrong. how do two column vectors multiply with each other}
\end{example}

\section{Moment system under Signal Temporal Logic (STL)} \label{sec:STLmoment}

% \red{this section should/ can be squeezed due to a) its standard stuff/ we don't contribute here, and 2) it is exceeding page limit: 6+n (https://2025.ieee-icra.org/contribute/final-paper-submission-instructions/)}

To describe a broad range of real-valued temporal properties commonly found in cyber-physical systems, a formal framework called Signal Temporal Logic (STL) is often used \cite{madsen2018metrics}.
For STL assessment purposes, a quantitative semantics to measure how well an STL formula $\varphi$ is satisfied by the signal $x(t)$, is provided by the robustness degree $\rho$.
% , \blue{as formalized in Theorem \ref{thm:Robustness} from the smooth robustness framework in \cite{pant2017smooth}.}
%, which is explained in the following theorem.
\begin{theorem} \label{thm:Robustness}
\cite{pant2017smooth} For a signal $x$ with domain $E$ s.t. $x : E \rightarrow \mathbb{R}$ and STL formula $\varphi$,  
if $\rho_{\varphi}(x, t) < 0$ then $\varphi$ is not satisfied by $x$ at time $t$,  
and if $\rho_{\varphi}(x, t) > 0$ then $x$ satisfies $\varphi$ at $t$. If $\rho_{\varphi}(x, t) = 0$ is inconclusive.
\end{theorem}
% \red{if a theorem is from the literature, should cite}
\begin{comment}
Similar to the STL semantics in \eqref{RecursiveSemantics}, $\rho$ is recursively defined as follows: 
{\small\begin{equation}
\begin{aligned}
&\rho^{\mathcal{\varpi}}(x,t) = h(x)-C_{\varphi}\\
&\rho^{\sim \varphi}(x,t) =  -\rho^{\varphi}(x,t)\\
&\rho^{\varphi_{1}\wedge \varphi_{2}}(x,t)=\min\left(\rho^{\varphi_{1}}(x,t),\rho^{\varphi_{2}}(x,t)\right)\\
&\rho^{\varphi_{1} \vee \varphi_{2}}(x,t)=\max\left(\rho^{\varphi_{1}}(x,t),\rho^{\varphi_{2}}(x,t)\right)\\
&\rho^{\mathsf{F}_{[t_i,t_f]} \varphi}(x,t)=\max_{t'\in\left[t+t_i,t+t_f\right]}\left( \rho^{\varphi}(x,t)  \right) \hspace{6em}\\ 
&\rho^{\mathsf{G}_{[t_i,t_f]} \varphi}(x,t)=\min_{t'\in \left[t+t_i,t+t_f\right]}\left( \rho^{\varphi}(x,t') \right)\\
&\rho^{\varphi_1 \Rightarrow \varphi_2}(x,t)=\max\left(-\rho^{\varphi_{1}}(x,t),\rho^{\varphi_{2}}(x,t')\right)
\label{RecursiveRho}
\end{aligned}
\end{equation}}
\end{comment}
\begin{example}
An ensemble of unicycles are required to visit two way-points and avoid an area along their trajectories over a time horizon $[t_0,t_N]$. The first way-point is visited within the time window $[t_a,t_b]$ and the second way-point is visited within $[t_c,t_N]$. The time window for the zone to avoid lies over the entire horizon, i.e., $[t_0,t_N]$. %Figure \ref{fig:ExTrajectoryRhoWPAvoid} shows the behavior of the ensemble trajectories and the compliance of the described task. 
\begin{comment}
\begin{figure}[h]
	\centering
    \includegraphics[width=0.45\textwidth] {figures/TrajectoryRhoWPAvoidV6.png}
		\caption{Way-points visiting and obstacle avoidance.
        \red{can remove if space is really tight}
        } 
	\label{fig:ExTrajectoryRhoWPAvoid}
\end{figure} 
\end{comment}
In accordance with the STL semantics in \cite{yang2020continuous}, the following formula only denotes the way-point visiting, considering that each way-point is eventually visited within its respective time-widow: 
\begin{equation}
\begin{aligned}
\varphi(m)=&\mathsf{F}_{[t_a,t_b]}(a^{\left<1\right>} m_{k1} + b^{\left<1\right>} m_{k2} - M^{\left<1\right>}_L \geq 0) \wedge \\
&\mathsf{F}_{[t_a,t_b]}(-a^{\left<1\right>} m_{k1} - b^{\left<1\right>} m_{k2} + M^{\left<1\right>}_U \geq 0)
\wedge \\
&\mathsf{F}_{[t_c,t_N]}(a^{\left<2\right>} m_{k1} + b^{\left<2\right>} m_{k2} - M^{\left<2\right>}_L \geq 0)
\wedge \\
&\mathsf{F}_{[t_c,t_N]}(-a^{\left<2\right>} m_{k1} - b^{\left<2\right>} m_{k2} + M^{\left<2\right>}_U \geq 0),
\end{aligned}
\label{eqVisitWP}
\end{equation}
% \red{explain that this comes from equation 13}
where the super-indices $\left<1\right>$ and $\left<2\right>$ are the first and second way-point, respectively. Expression \eqref{eqVisitWP} is in moment space representation, following the formulation in \eqref{eq:polyhedron_moment}. Then, the transformation into the robustness degree considers the STL semantics using the $\max$ and $\min$ operators, which are approximated by the Log-Sum-Exponential (LSE) form \cite{mao2022successive}.
\begin{comment}
, as shown below:  
\begin{equation}
\begin{aligned}
\widetilde{\max}\left[ \left( a_{1},..., a_{n} \right) \right]^{\top}:= \frac{1}{K}\log\left( \sum_{i=1}^{n}e^{Ka_{i}} \right) 
\label{eq:maxLSE}
\end{aligned}
\end{equation}

\begin{equation}
\begin{aligned}
\widetilde{\min}\left[ \left( a_{1},..., a_{m} \right) \right]^{\top}:= -\frac{1}{K}\log\left( \sum_{i=1}^{n}e^{-Ka_{i}} \right)
\label{eq:minLSE}
\end{aligned}
\end{equation}

The $\max$ and $\min$ operators approximation in \eqref{eq:maxLSE} and \eqref{eq:minLSE} respectively, is smooth when using LSE expressions. As the parameter $K \rightarrow \infty$, the maximum and minimum values approach to their true values. 
\end{comment}
\end{example}
Considering the previous formula semantics and smooth LSE approximations to find the robustness degree $\rho(m)$ in moment space, the following mixed integer nonlinear optimal control problem is presented: 
\begin{comment}
\begin{equation}
\begin{split}
&\max_{m_k, u, z_{i}} \underbrace{\rho(m)}_{\textbf{Robustness degree}} - \underbrace {\delta\|m_k(T) - m_k(t_f)\|_2^2}_{\textbf{Final reference}} \\  
& \hspace{9em} - \underbrace{\alpha \left[u^{\top}u + v^{\top}v\right]}_{\textbf{LQR term}}
\\
&\text{s.t.} \\ & \dot{m_k} = \left(u \hat{B_1} + v \hat{B_2} \right)m_k + \left(u \Tilde{B_3} + v \Tilde{B_4}\right)m_k, \\
&\textbf{Set of equations in \eqref{eq:setPolyObsMspace}},\\
&\sum_{i=1}^d z_i = 1\end{split}
\label{eq:optimalControlWPObs}
\end{equation} 
\end{comment}
\begin{equation}
\begin{aligned}
    &\max_{m,u,v:[0,T]\rightarrow\mathbb{R}} \hspace{0.1em} 
        \rho(m) - \|\tilde m(T) - \tilde m_F\|^2 
          - \left(u^{\top}u + v^{\top}v\right) \\
   \text{s.t.}& \quad 
        \frac{d}{dt} m(t) 
            = u(t)\mathcal{B}_1 m(t) + v(t)\mathcal{B}_2 m(t), \\
        & -(I\otimes M)(m^+\otimes \mathbf{1})
          - (I\otimes(b_i+M))(m^-\otimes\mathbf{1}) \\
        & \leq (I\otimes a_i)m(t)z_i \\
        & \leq (I\otimes(b_i+M))(m^+\otimes\mathbf{1})
          + (I\otimes M)(m^+\otimes \mathbf{1}). \\
        & \forall i=1, \cdots,d  
\end{aligned}
\label{eq:WayPointObsAvoid}
\end{equation}
% \red{re-check after reviewing the formulation in the constraint case}
%where the decision variables are the moment $m_k$, the control signals $u$ and binary variables $z_i$.
%For clarity of notation, the time dependence of the decision variables has been omitted, although it remains present throughout the formulation.
Notice that it is pursued to maximize the objective function to obtain a positive robustness degree, meaning that the resulting trajectory is in compliance with the task specifications in the formula. The obstacle avoidance approach described in section \ref{ObsAvoid} is included as a set of hard constraints. Additionally, two penalization terms are included: the Euclidean norm between the final state and the desired state, and the LQR term to avoid large values in the control signals.

\begin{comment}
\section{Analysis}

\red{what do possibly prove:
1) Reachability analysis (corresponding to controllability in the unconstrained case)
2) The constraint in the truncated moment space will converge (to the exact one) as the order increases. This is also empirically verified by comparing \Cref{UnicycleConsNtrun4} and \Cref{UnicycleConsNtrun8}
}

\section{PLANNING AND CONTROL IN MOMENT SPACE}
with truncation, design a terrain-agnostic controller that can work fairly well on all different terrains.
\end{comment}
\begin{figure}[t]
    \centering
    % Each subfigure takes ~47% of column width to fit nicely side by side
    \begin{subfigure}{0.49\linewidth}
        \centering
        \includegraphics[width=\linewidth]{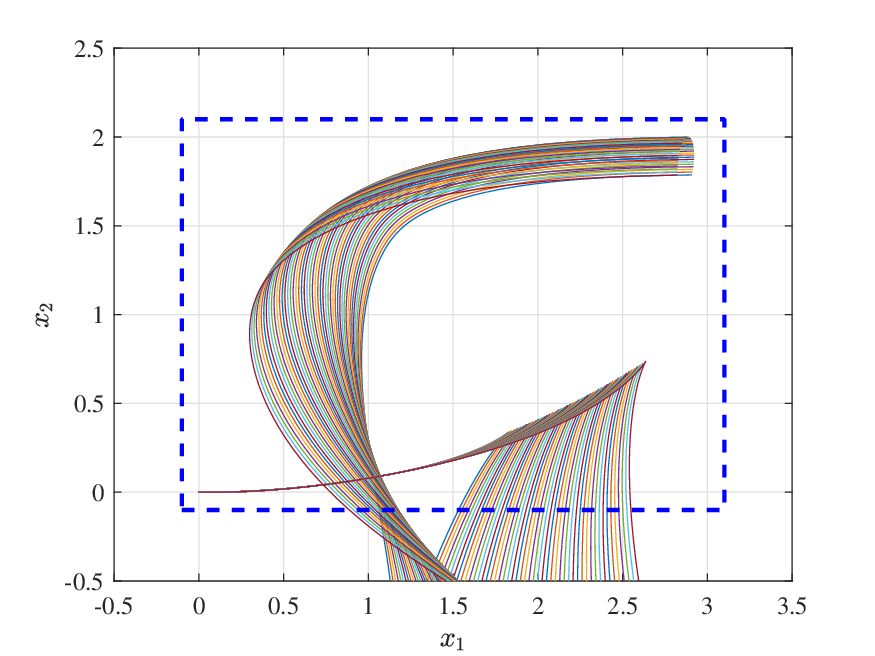}
        \caption{No box constraint.}
        \label{FigNoBoxEx4}
    \end{subfigure}
    \hfill
    \begin{subfigure}{0.49\linewidth}
        \centering
        \includegraphics[width=\linewidth]{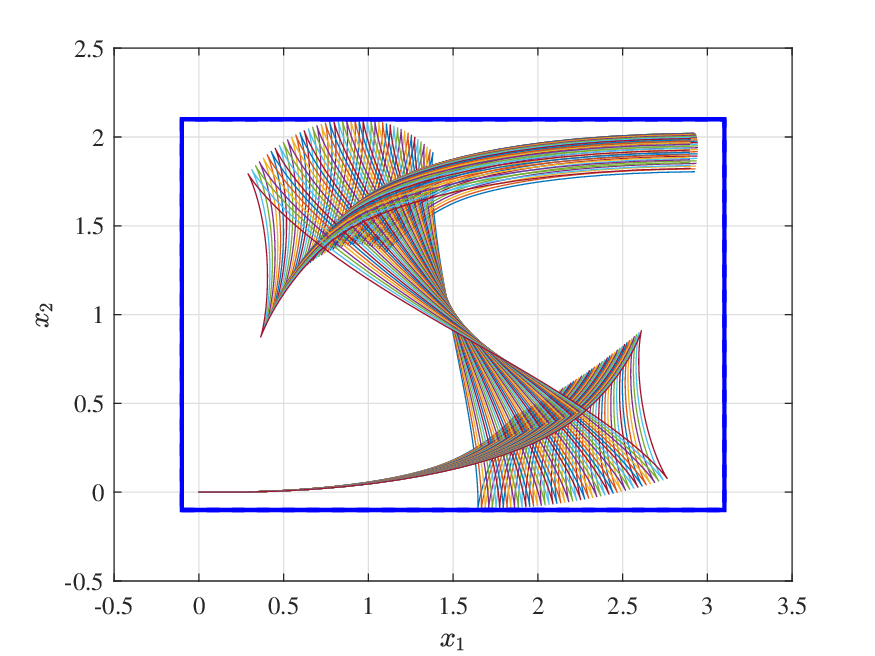}
        \caption{Box-constrained trajectories.}
        \label{FigBoxEx4}
    \end{subfigure}
    \caption{Comparison between unconstrained and box-constrained trajectories.}
    \label{fig:comparisonBox}
\end{figure}

\begin{figure}[t]
    \centering
    \begin{subfigure}{0.23\textwidth}
        \centering
        \includegraphics[width=\linewidth]{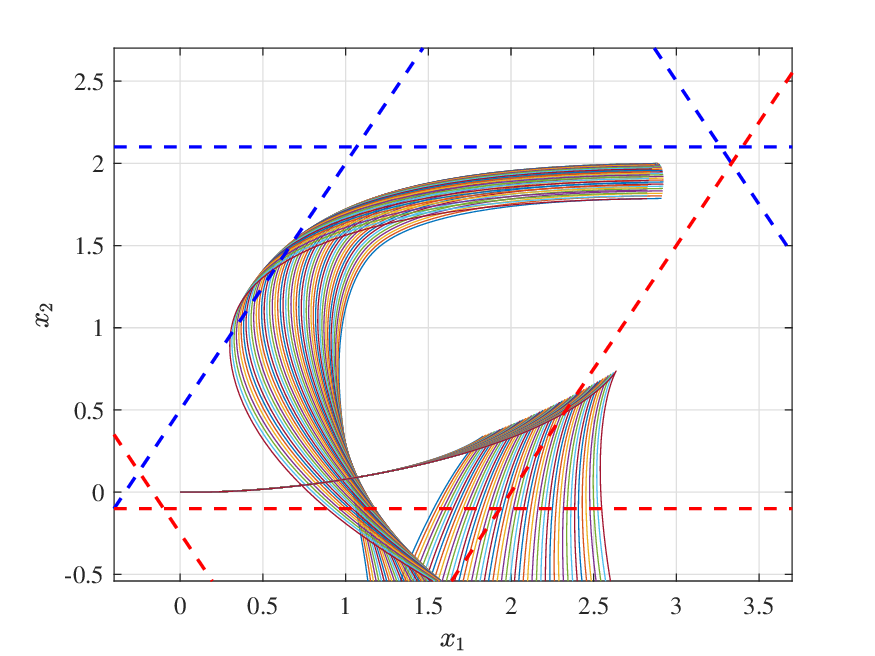}
        \caption{No polyhedron constraint.}
        \label{UnicycleUncons}
    \end{subfigure}
    \hspace{0.005\textwidth} % smaller gap
    \begin{subfigure}{0.23\textwidth}
        \centering
        \includegraphics[width=\linewidth]{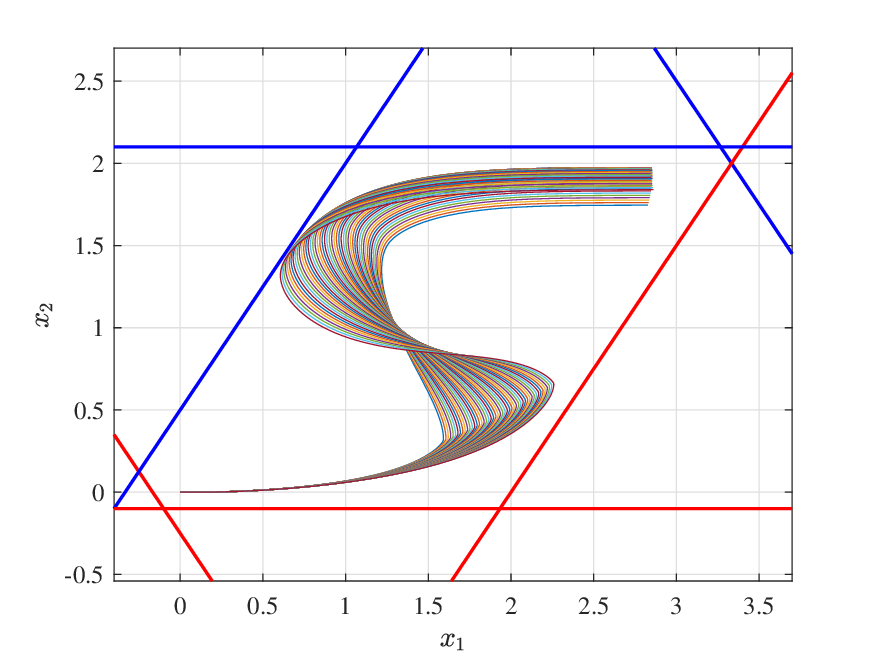}
        \caption{Polyhedron-constrained trajectories.}
        \label{UnicycleConsNtrun8}
    \end{subfigure}
    \caption{Comparison between unconstrained and polyhedron-constrained trajectories.
    }
    \label{fig:comparisonPoly}
\end{figure}

\section{SIMULATION RESULTS} \label{sec:simulation}

To demonstrate the proposed framework, our constrained ensemble control is applied to the unicycle model in \eqref{eq:bicycle}, considering the way-point visiting and obstacle avoidance approaches. %The optimal control problems for the box constraint and polyhedron constraint are solved using the $fmincon$ function with the interior-point solver in Matlab. 
The ensemble control is performed with $0.9 \leq \eta \leq 1.1$. The optimal control problem is solved using General Algebraic Modeling System GAMS with IPOPT (NLP problem) and DICOPT (MINLP problem) solvers \cite{DICOPT2024}. All the experiments were carried out on a machine equipped with an Intel Core Ultra 7 (1.4 GHz, 16 cores, 22 logical threads) and 16 GB of RAM.

\begin{figure*}[t]
    \centering

    % ---------- Row 1 ----------
    \begin{subfigure}{0.49\textwidth}
        \centering
        \includegraphics[width=0.48\textwidth]{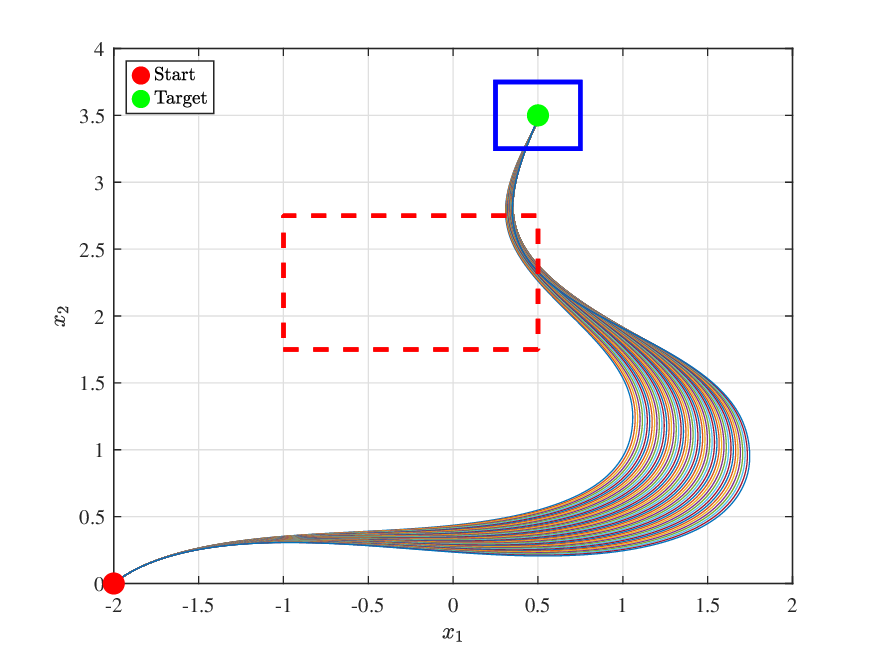}
        \includegraphics[width=0.48\textwidth]{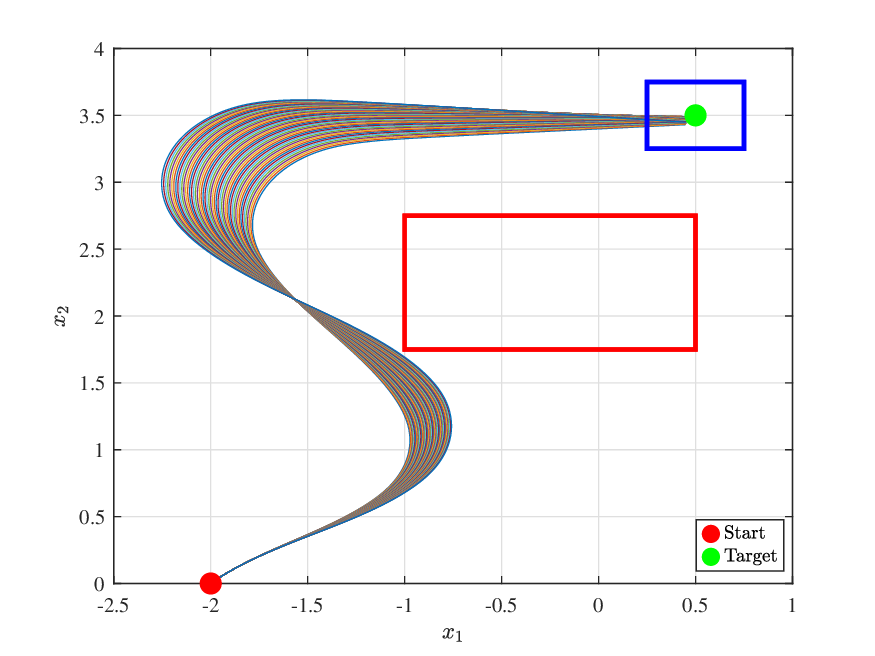}
        \caption{One-obstacle environment (Offline simulation).}
        \label{fig:OneObs}
    \end{subfigure}
    \hfill
    \begin{subfigure}{0.49\textwidth}
        \centering
        \includegraphics[width=0.48\textwidth]{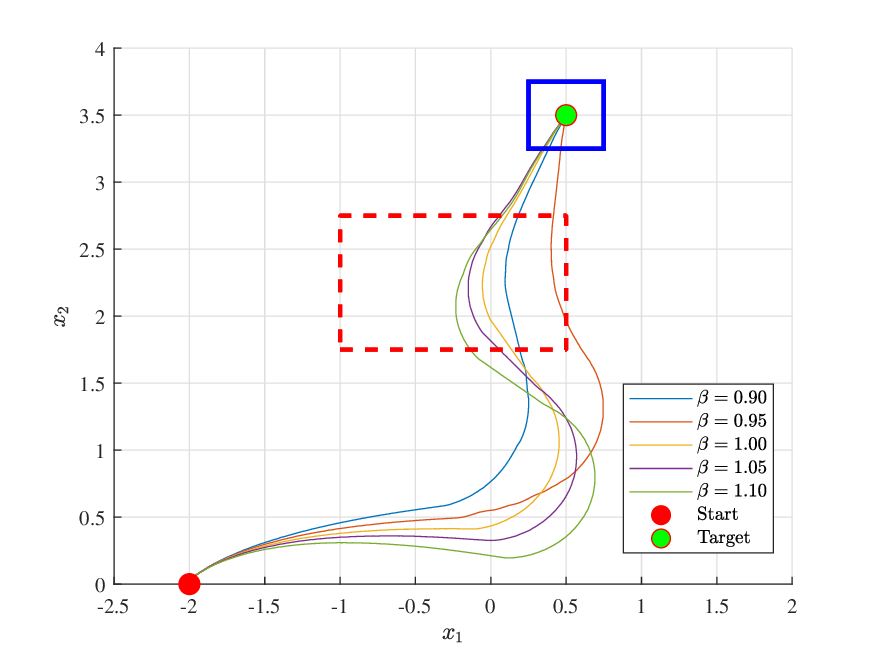}
        \includegraphics[width=0.48\textwidth]{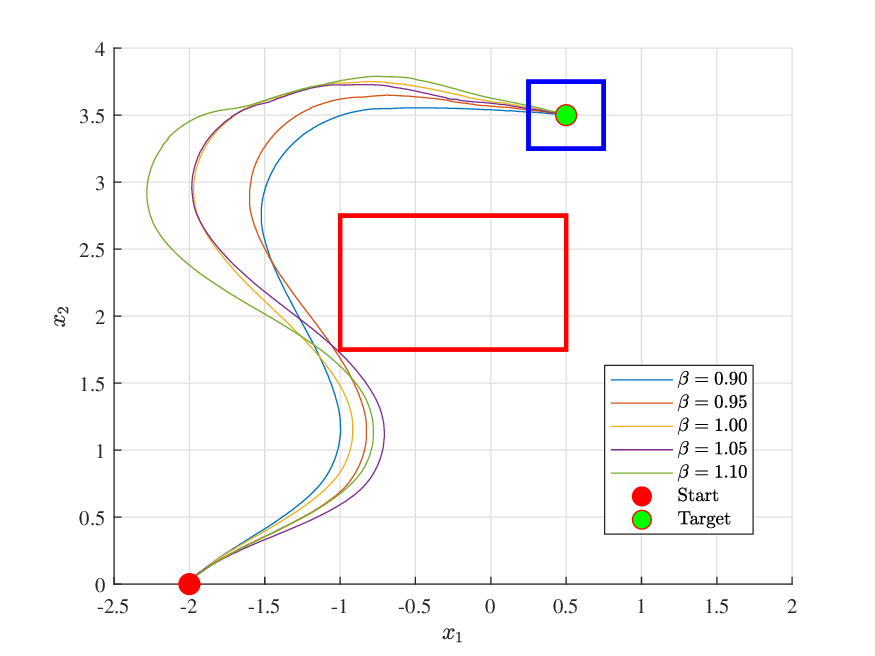}
        \caption{One-obstacle environment (Online hardware experiment).}
        \label{fig:OneObsHard}
    \end{subfigure}

    \vspace{0.4cm}

    % ---------- Row 2 ----------
    \begin{subfigure}{0.49\textwidth}
        \centering
        \includegraphics[width=0.49\textwidth]{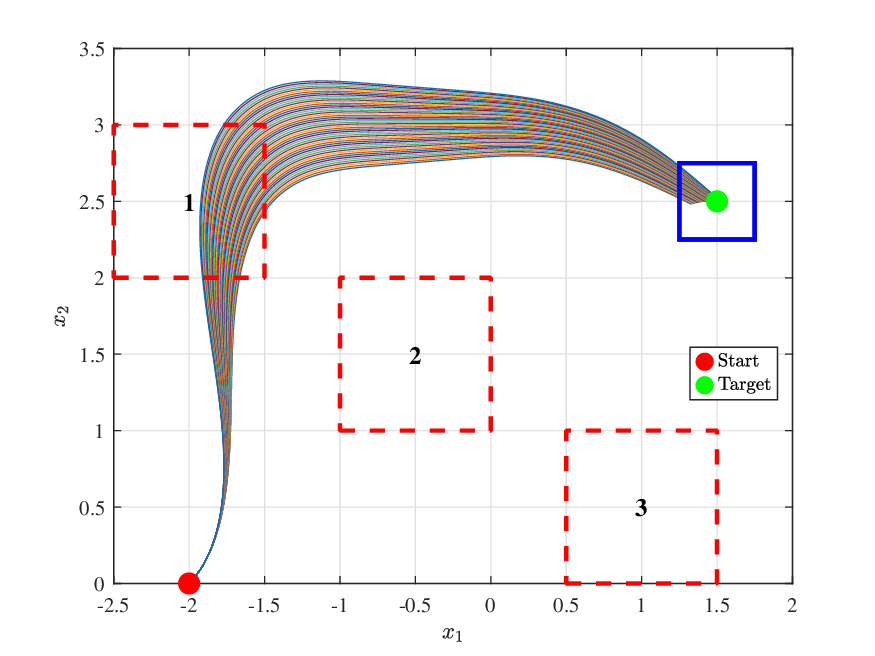}
        \includegraphics[width=0.49\textwidth]{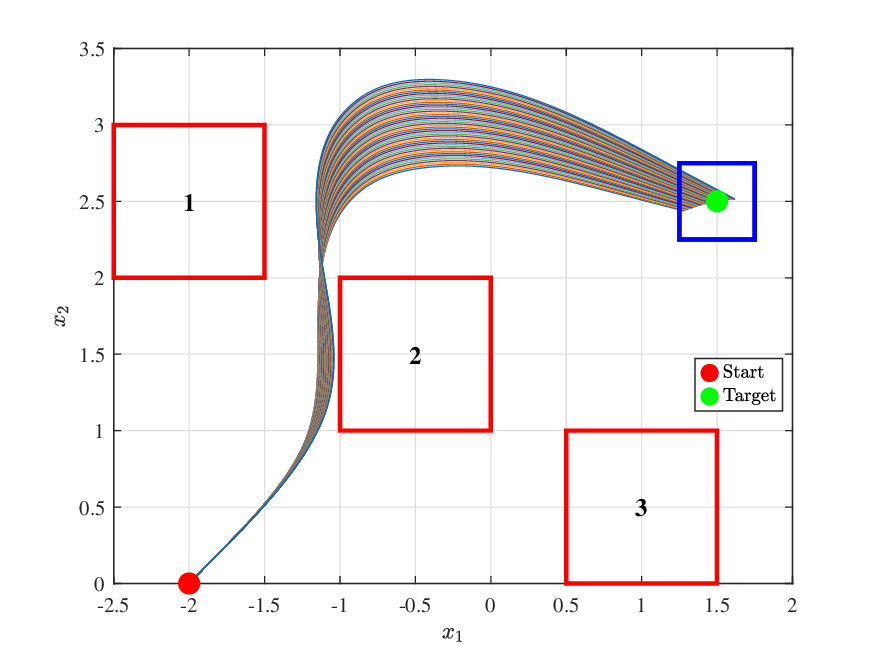}
        \caption{Three-obstacle environment (Offline simulation).}
        \label{fig:ThreeObs}
    \end{subfigure}
    \hfill
    \begin{subfigure}{0.49\textwidth}
        \centering
        \includegraphics[width=0.48\textwidth]{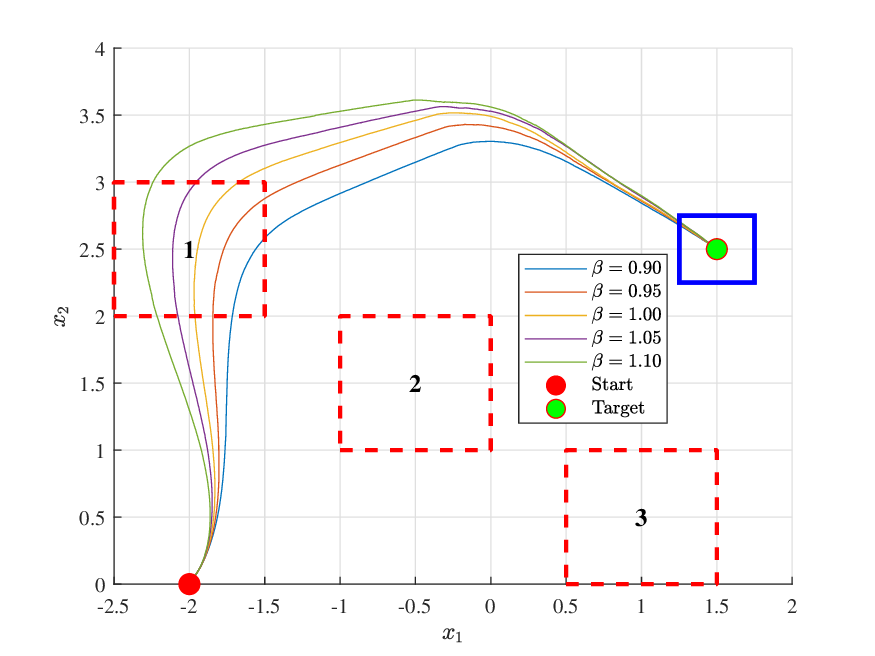}
        \includegraphics[width=0.48\textwidth]{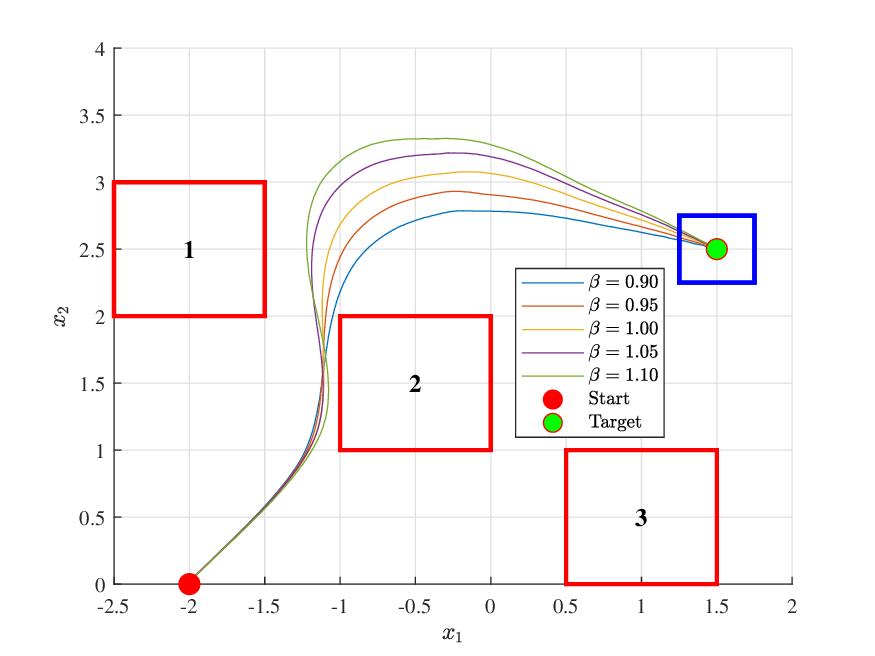}
        \caption{Three-obstacle environment (Online hardware experiment).}
        \label{fig:ThreeObsHard}
    \end{subfigure}

    \vspace{0.4cm}

    % ---------- Row 3 ----------
    \begin{subfigure}{0.49\textwidth}
        \centering
        \includegraphics[width=0.48\textwidth]{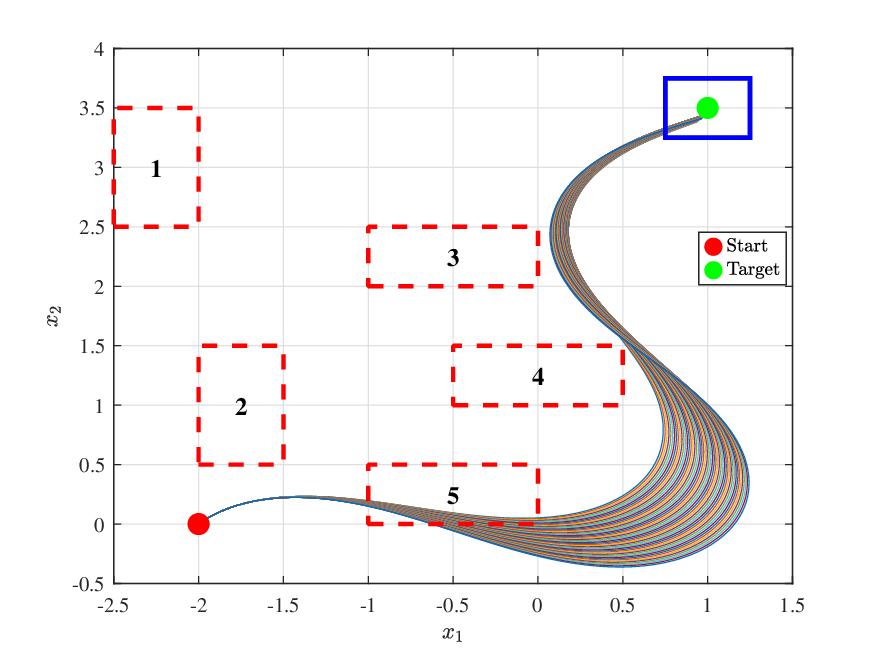}
        \includegraphics[width=0.48\textwidth]{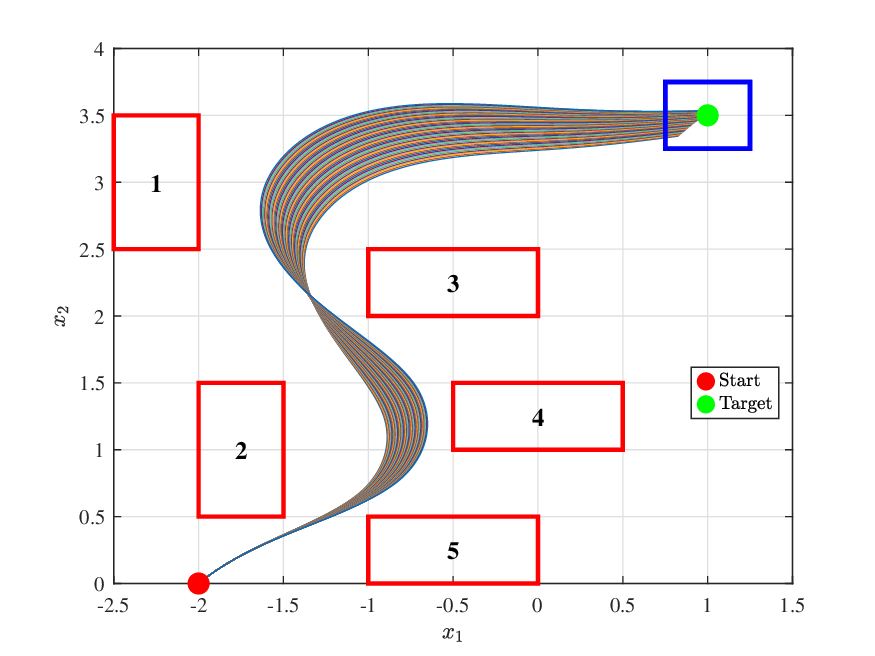}
        \caption{Five-obstacle environment (Offline simulation).}
        \label{fig:FiveObs}
    \end{subfigure}
    \hfill
    \begin{subfigure}{0.49\textwidth}
        \centering
        \includegraphics[width=0.48\textwidth]{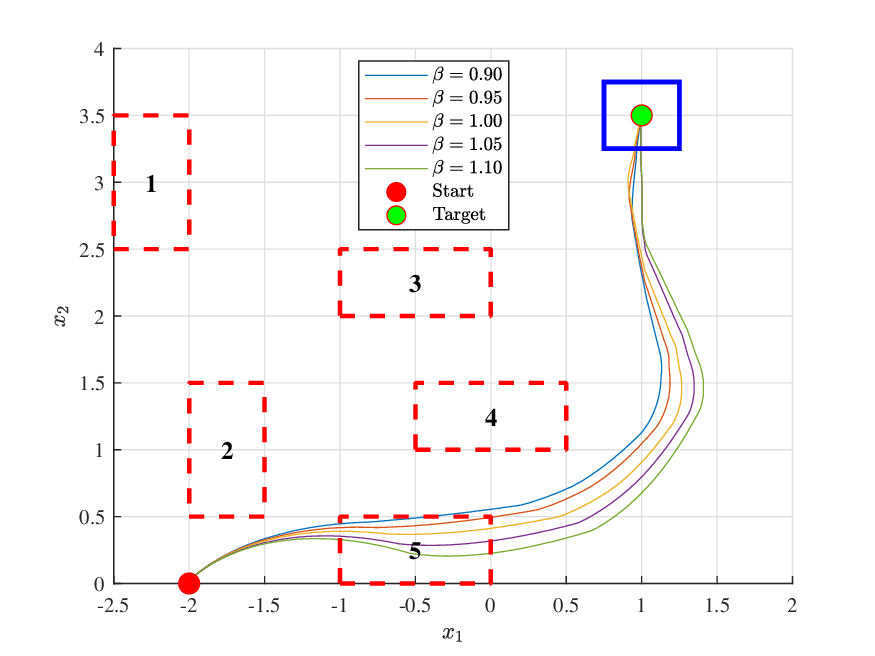}
        \includegraphics[width=0.48\textwidth]{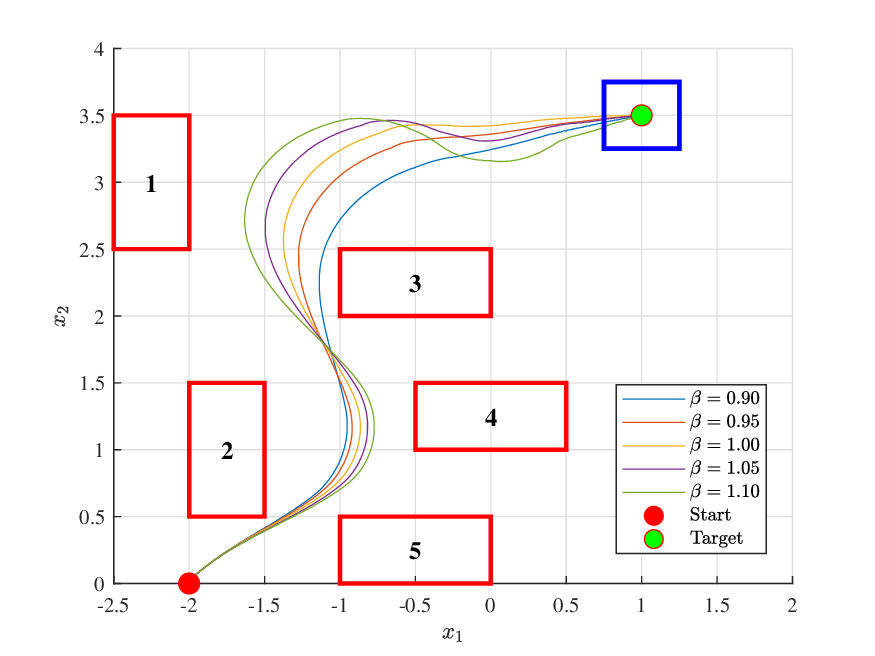}
        \caption{Five-obstacle environment (Online hardware experiment).}
        \label{fig:FiveObsHard}
    \end{subfigure}

    % ---------- Main Caption ----------
    \caption{Final way-point visiting and obstacle avoidance for different environment configurations. The blue square around the target indicates the way-point boundaries. An $8\text{th}$ truncation order for all the experiments is used, considering the unconstrained (dash-line, inactive obstacles) and the constrained settings (solid-line, active obstacles), for offline simulation (columns 1 and 2) and hardware experiment (columns 3 and 4)}
    \label{fig:offlineAvoid}
\end{figure*}

\subsection{Steering an ensemble of unicycles within a box constraint}

For the box constraint case, it is desired to find a sequence of control signals that drives the trajectories of the unicycles from a start position to a target position within a bounded region (box), where the initial and final positions are in $[0,0]$ and $[3,2]$ respectively, the time horizon is 2 s and the truncation order is 4. For simulation purposes, the time step is 0.01 s. Figure \ref{FigNoBoxEx4} presents the ensemble trajectories with no box constraint. Once the box constraint is active, the trajectories are effectively driven within the bounded region as shown in Figure \ref{FigBoxEx4}.

\subsection{Steering an ensemble of unicycles within a polyhedron constraint}

Given the set of equations in \eqref{eq:SetPoly}  that represents a polyhedron in $\mathbb{R^2}$, an ensemble of unicycles is driven from $[0,0]$ to $[3,2]$ in a time horizon of 2 s.
\begin{equation} \label{eq:SetPoly}
    \begin{split}
        &-0.5<3x_1+2x_2<14\\
        &-6<-3x_1+2x_2<1 \\
        &-0.1<x_2<2.1
    \end{split}
\end{equation}
In the absence of constraints, the trajectories are free to move throughout the space, extending beyond the boundaries of the polyhedron before reaching the target, as shown in Figure \ref{UnicycleUncons}. Once the constraints are included with an $8^{\text{th}}$ truncation order, the trajectories become smooth and remain inside the bounded region as presented in Figure \ref{UnicycleConsNtrun8}.

%Notice that the trajectories are not smooth and a tiny portion are out of the polyhedron. Furthermore, if the truncation order is increased to 8, the trajectories are smoother and fully enclosed within the bounded region, as illustrated in Figure \ref{UnicycleConsNtrun8}.

\begin{comment}
\begin{figure*}[t]
    \centering
    \begin{subfigure}{0.32\textwidth}
        \centering
        \includegraphics[width=\linewidth]{figures/NoPolyConstEx4.png}
        \caption{No polyhedron constraint.}
        \label{UnicycleUncons}
    \end{subfigure}
    \hspace{0.005\textwidth} % smaller gap
    \begin{subfigure}{0.32\textwidth}
        \centering
        \includegraphics[width=\linewidth]{figures/PolyhedronConstraintNtrun4.png}
        \caption{Polyhedron constraint (Truncation order of 4).}
        \label{UnicycleConsNtrun4}
    \end{subfigure}
    \hspace{0.005\textwidth} % smaller gap
    \begin{subfigure}{0.32\textwidth}
        \centering
        \includegraphics[width=\linewidth]{figures/PolyhedronConstraintNtrun8.png}
        \caption{Polyhedron constraint (Truncation order of 8).}
        \label{UnicycleConsNtrun8}
    \end{subfigure}
    \caption{Comparison of unicycle trajectories with and without polyhedron constraints for different truncation levels.}
    \label{fig:comparisonPoly}
\end{figure*}
\end{comment}

\subsection{Way-point visit and obstacle avoidance}

\begin{figure}[t]
    % ---------- Row 1 ----------
    \begin{subfigure}{0.48\textwidth}
        \centering
        \includegraphics[width=0.48\textwidth]{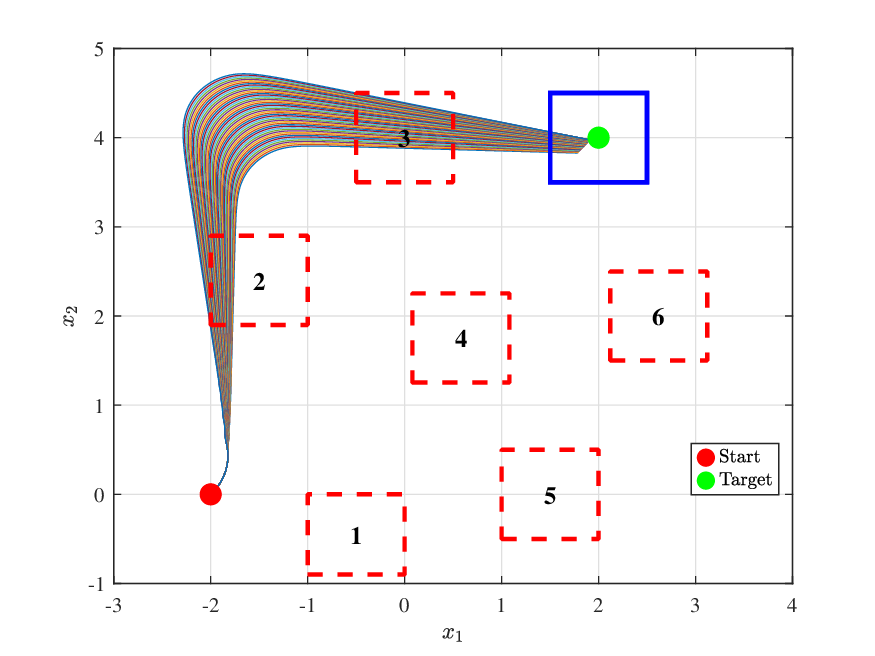}
        \includegraphics[width=0.48\textwidth]{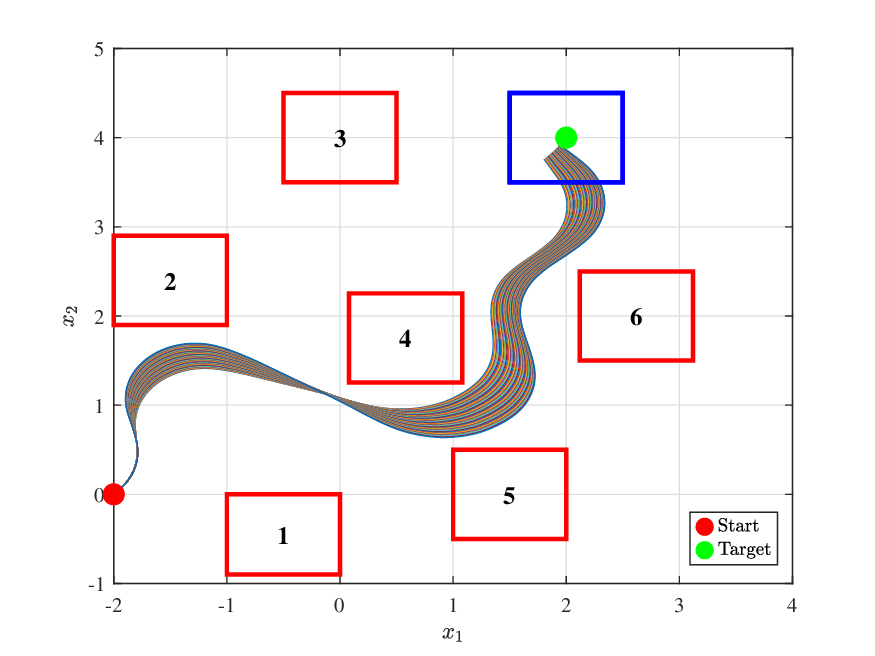}
        \caption{Six-obstacle environment.}
        \label{fig:pair1}
    \end{subfigure}
    
    \vspace{0.4cm}

    % ---------- Row 2 ----------
    \begin{subfigure}{0.48\textwidth}
        \centering
        \includegraphics[width=0.48\textwidth]{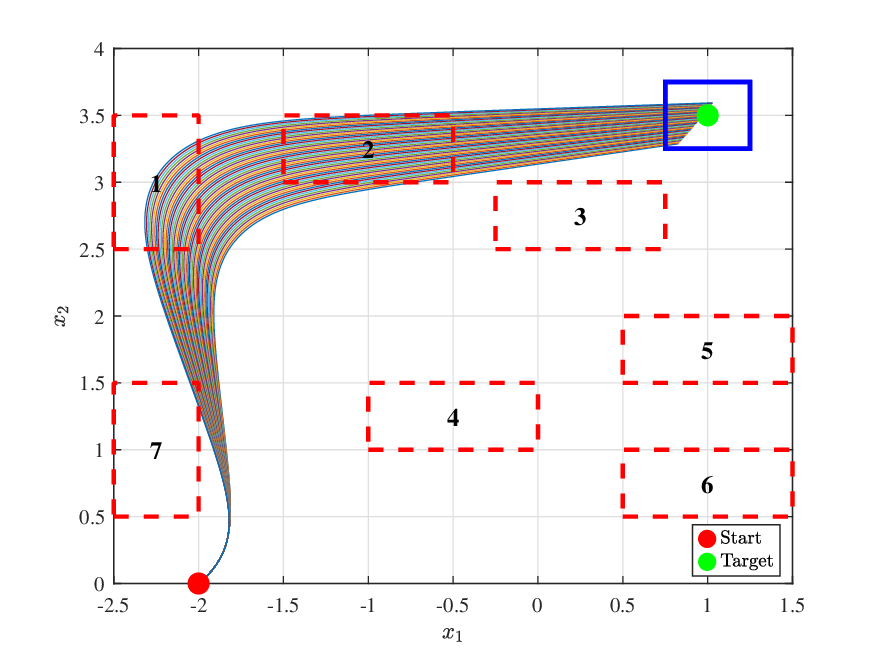}
        \includegraphics[width=0.48\textwidth]{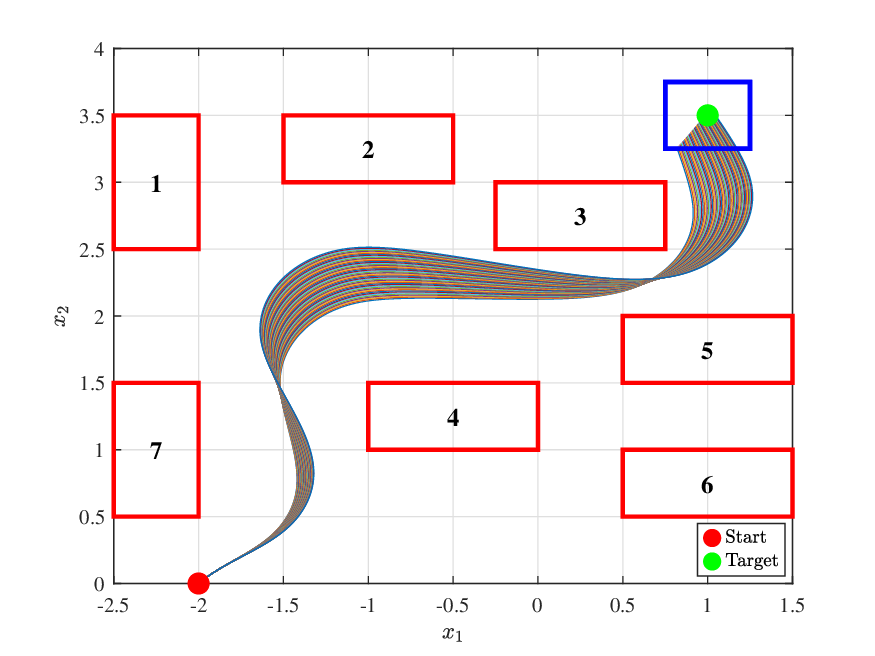}
        \caption{Seven-obstacle environment.}
        \label{fig:pair3}
    \end{subfigure}
    
    \vspace{0.4cm}

    % ---------- Row 3 ----------
    \begin{subfigure}{0.48\textwidth}
        \centering
        \includegraphics[width=0.48\textwidth]{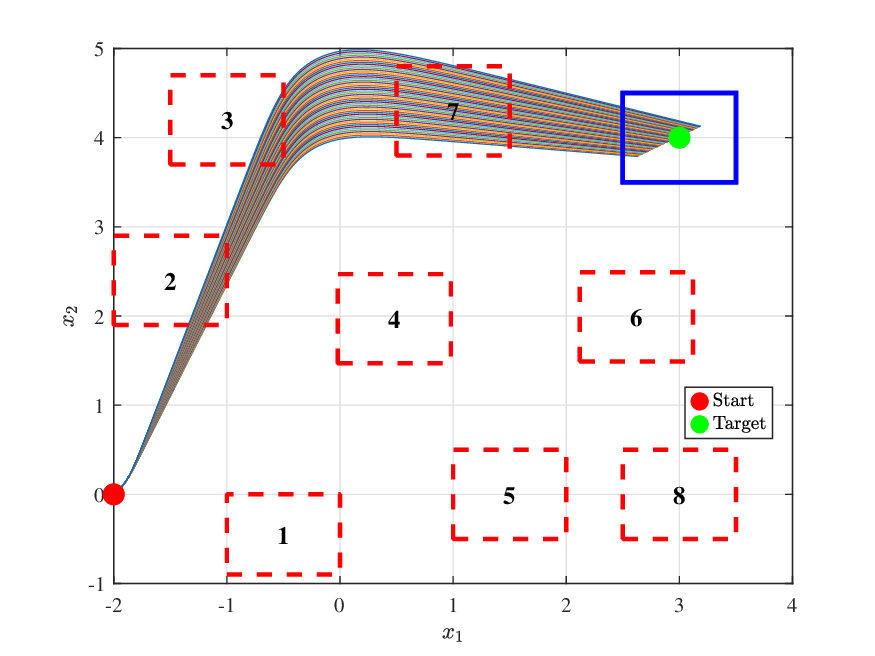}
        \includegraphics[width=0.48\textwidth]{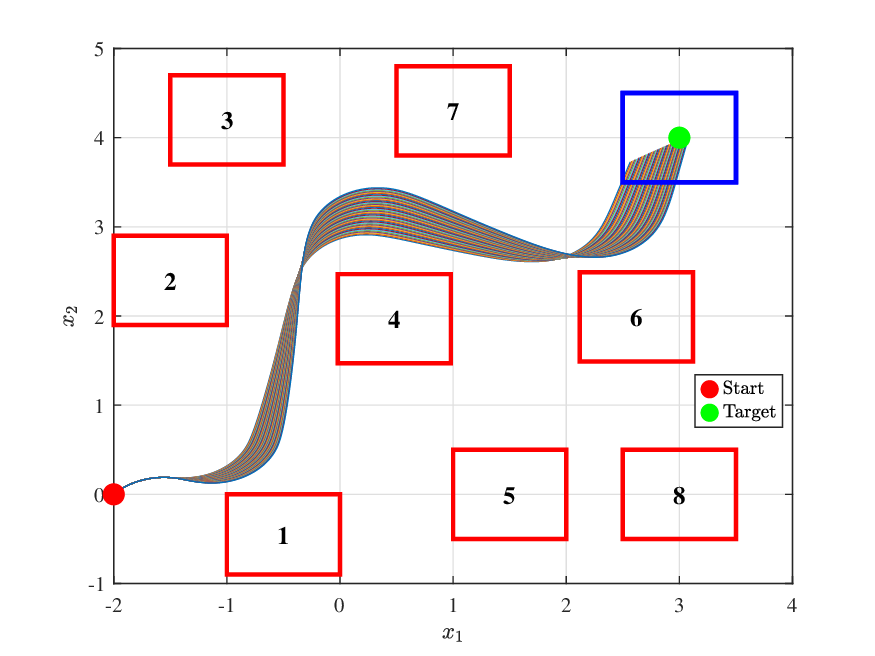}
            \caption{Eight-obstacle environment.}
        \label{fig:pair5}
    \end{subfigure}
    
    % ---------- Main Caption ----------
    \caption{Simulations with more complex scenarios.}
    \label{fig:SimulationMoreComplex}
\end{figure}

This approach is addressed using the formulation in \eqref{eq:WayPointObsAvoid}. In this case, the task for the ensemble of trajectories is to reach a final way-point within the last two seconds of a time horizon of $16 s$, while avoiding obstacles. Figures \ref{fig:OneObs}, \ref{fig:ThreeObs}, and \ref{fig:FiveObs} present the ensemble trajectories of the unicycles for one, three and five obstacles, respectively. In the unconstrained setup, i.e., absence of obstacles, the aim only lies on arriving the final way-point in the specified time window, and the trajectories freely move around the space. In the constrained settings, the obstacles are effectively avoided. Environments with more obstacles, are shown in Figure \ref{fig:SimulationMoreComplex}.

%and always avoid an obstacle. The truncation order is increased to 8 Figure \ref{fig:TrajectoryRhoWP} presents the way-point visiting of the trajectories with start and target positions of $[1,1]$ and $[6.5,5.5]$, respectively. Notice in Figure \ref{fig:TrajectoryRhoWPAvoid} that the trajectories effectively avoid the obstacle once this is included along their path.   

\begin{comment}
\begin{figure*}[t]
    \centering
    \begin{subfigure}{0.49\textwidth}
        \centering
        \includegraphics[width=\linewidth]{figures/WPVisit.png}
        \caption{Way-points visiting.}
        \label{fig:TrajectoryRhoWP}
    \end{subfigure}
    \hspace{0.005\textwidth} % smaller gap
    \begin{subfigure}{0.49\textwidth}
        \centering
        \includegraphics[width=\linewidth]{figures/WPandObs.png}
        \caption{Way-points visiting and obstacle avoidance.}
        \label{fig:TrajectoryRhoWPAvoid}
    \end{subfigure}
    \hspace{0.005\textwidth} % smaller gap
    \caption{Unicycle trajectories for the way-point visiting and obstacle avoidance. }
    \label{fig:comparisonPoly}
\end{figure*}
\end{comment}

\section{Hardware Experiments} \label{sec:hardwareExp}
\begin{figure}[t]
    \centering
    
    % ---------- Left image ----------
    \begin{subfigure}{0.22\textwidth}
        \centering
        \includegraphics[width=\textwidth]{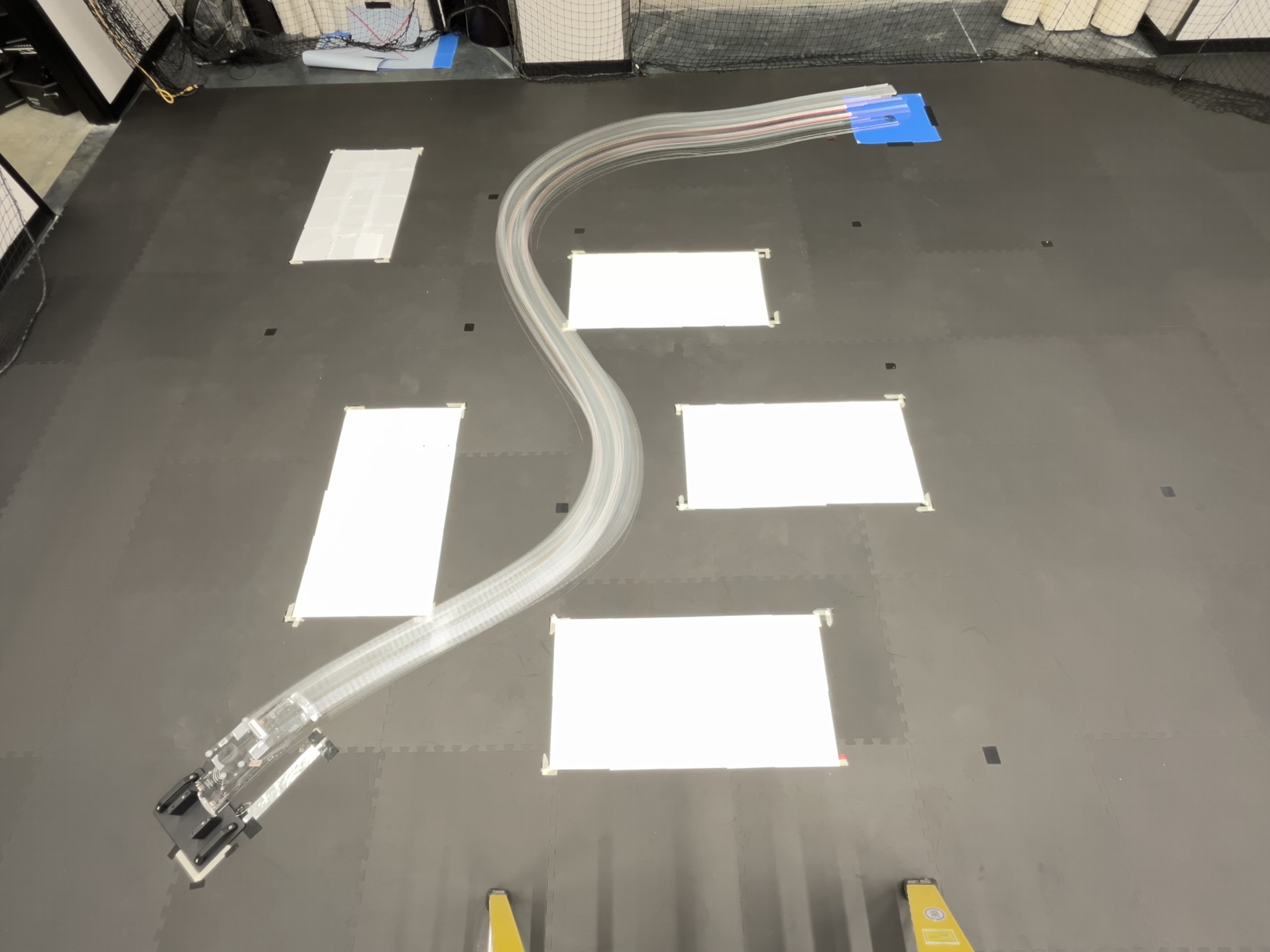}
        \caption{Space setup for test trajectories.}
        \label{fig:LabQcar}
    \end{subfigure}
    \hfill
    % ---------- Right image ----------
    \begin{subfigure}{0.238\textwidth}
        \centering
        \includegraphics[width=\textwidth]{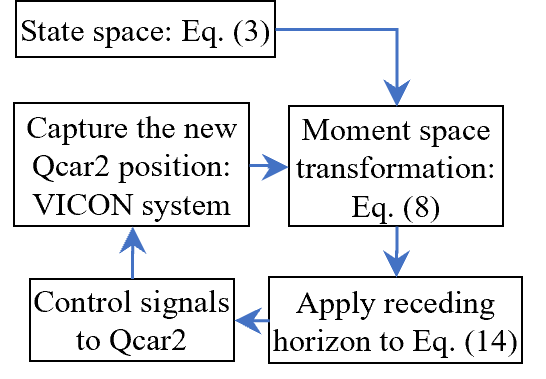}
        \caption{Online architecture.}
        \label{fig:MPCOnline}
    \end{subfigure}

    \caption{QCar 2 hardware settings.
    %\red{adjust the width of the boxes to be the same}
    %\red{equation numbers might need to be changed after Wei tailors the unconstrained section}
    }
    \label{fig:HardwareExperiments}
\end{figure}

For validation purposes in real scenarios, the constrained ensemble control approach is tested using the Qcar 2, a $1/10^{\text{th}}$ scale vehicle designed for academic self-driving initiatives. Figure \ref{fig:LabQcar} presents the experimental setup in the VICON camera-equipped laboratory.
\begin{comment}
\begin{figure}[t]
    \centering
    
    % ---------- Left image ----------
    \begin{subfigure}{0.22\textwidth}
        \centering
        \includegraphics[width=\textwidth]{figures/LabQcar2.jpg}
        \caption{Space setup for test trajectories.}
        \label{fig:LabQcar}
    \end{subfigure}
    \hfill
    % ---------- Right image ----------
    \begin{subfigure}{0.238\textwidth}
        \centering
        \includegraphics[width=\textwidth]{figures/MPCOnline.png}
        \caption{Online architecture.}
        \label{fig:MPCOnline}
    \end{subfigure}

    \caption{QCar 2 hardware settings.
    \red{adjust the width of the boxes to be the same}
    \red{equation numbers might need to be changed after Wei tailors the unconstrained section}
    }
    \label{fig:HardwareExperiments}
\end{figure}
\end{comment}
Since the trajectories obtained in offline simulations do not match with the real scenario deployment, caused by the error propagation, we have implemented an online strategy that collects the current robot's position and runs the optimal control problem, prior to sending a predefined number of samples of the time-horizon to the Qcar 2. In this sense, the ground vehicle progressively performs the trajectory, fulfilling the requirements contained in the optimization problem. Figure \ref{fig:MPCOnline} presents the loop architecture in the online settings. The state space equations are transformed into a moment space representation. Then, under the receding horizon principle, successive optimal control problems are solved, based on the current Qcar 2 position along the trajectory. %Start and target positions are transformed into moment space, then the optimal control problem is solved for the entire horizon $N$, providing a sequence of control signals $u(t) \hspace{0.3em} \forall t \in [0,N_{0}]$, being $N_{0} < N$, moving the Qcar 2 to a position $x_1$. If the target has been reached, then the process is terminated; otherwise, the current position $x_{0}$ is updated and transformed in moment space to repeat the process.}
% \red{summarize as receding horizon control}
Experiments considering final way-point visiting and obstacle avoidance are shown in Figures \ref{fig:OneObsHard}, \ref{fig:ThreeObsHard} and \ref{fig:FiveObsHard}, for environments of one, three, and five obstacles, respectively. A better visualization of the trajectories is shown in the multimedia attachment.
The ensemble control is performed with $0.9 \leq \eta \leq 1.1$, and both the unconstrained and constrained settings are included for comparison purposes. As noted in the experiments, the trajectories are moving around the space in the unconstrained settings, regardless of the obstacles' position, whereas in the constrained settings, the obstacles are correspondingly avoided throughout the entire paths.

\section{CONCLUSIONS} \label{sec:conclusions}
% \red{constrained case should come before simulation; be concise, fairly high level; following the style in the abstract} 
We formulate a novel constrained ensemble control framework to steer an ensemble of unicycle vehicles in polyhedron-constrained and obstacle-populated environments. STL specifications are incorporated as soft constraints to encode task requirements. Moreover, the moment kernel-transformed obstacle avoidance formulations are included as a set of hard constraints within the optimal control problem. Simulations and hardware experimentation demonstrate the effectiveness of the approach by broadcasting a shared controller to satisfy task specifications in a constrained environment.

\bibliography{reference}
\bibliographystyle{IEEEtran}

\end{document}

%% file: preamble.tex
\usepackage{amsmath,amsthm,amssymb, amsfonts} %
\usepackage{bbold}
\usepackage{mathtools}
\usepackage{graphicx}
\usepackage{xcolor}

\usepackage[export]{adjustbox} %

\usepackage{pdfpages}
\usepackage{bm}
\usepackage{url}
\usepackage{enumerate}
\usepackage{float,balance}
\usepackage[sort, compress]{cite}
\usepackage{cases,balance}

\usepackage[capitalise]{cleveref}

\usepackage[font=footnotesize]{caption}

\usepackage{array}
\usepackage{booktabs}
\usepackage{colortbl}
\usepackage{multirow} %
\usepackage{threeparttable} %

\usepackage{placeins}
\usepackage{ulem}  %

\usepackage{comment} %
\excludecomment{comments} %

\usepackage{algorithm}

\usepackage{tikz}

\newcommand{\setParam}{\boldsymbol{\Psi}}

\newcommand{\setX}{\mathbf{X}} %

\newcommand{\R}{\mathbb{R}}

\newcommand{\red}[1]{{\color{red}#1}}

\newcommand{\orange}[1]{{\color{orange}#1}}

\newcommand{\edit}[1]{\textcolor{black}{#1}}

\newcommand{\tr}{^\top}

\theoremstyle{plain}%

\newtheorem{theorem}{Theorem}

\theoremstyle{remark}

\theoremstyle{definition}
\newtheorem{example}{Example}

\graphicspath{{figures/}}